\documentclass[poms,final,nonblindrev]{poms1_V1} 

\OneAndAHalfSpacedXI 

\usepackage{graphicx}
\usepackage{subfigure, epsfig}

\usepackage{amsmath,amssymb,amsfonts}
\usepackage{algorithmic}
\usepackage{graphicx}
\usepackage{textcomp}
\usepackage{xcolor}
\usepackage{tcolorbox}
\usepackage[style=authoryear,
            backend=biber,
            maxcitenames=2, 
            mincitenames=1,  
            uniquename=false, 
            uniquelist=false  
            ]{biblatex}
\TheoremsNumberedThrough     
\ECRepeatTheorems

\EquationsNumberedThrough    

\usepackage{fancyhdr} 

\begin{document}


\RUNAUTHOR{Yin et al.}

\RUNTITLE{Rethinking Supply Chain Planning}

\TITLE{
Rethinking Supply Chain Planning: A Generative Paradigm
}


\ARTICLEAUTHORS{
\begin{center}
    \setlength{\parskip}{0pt}
    
    \normalsize \bfseries 
    Jiaheng Yin\textsuperscript{1,2}, 
    Yongzhi Qi\textsuperscript{2,*}, 
    Jianshen Zhang\textsuperscript{2}, 
    Dongyang Geng\textsuperscript{2}, 
    Zhengyu Chen\textsuperscript{2}, 
    Hao Hu\textsuperscript{2}, \\
    Wei Qi\textsuperscript{1,*}, 
    Zuo-Jun Max Shen\textsuperscript{3,4}
    \\

    \small \normalfont \itshape
    \textsuperscript{1}Tsinghua University, Beijing, China;
    \textsuperscript{2}JD.com, Beijing, China;
    \textsuperscript{3}Faculty of Engineering and Faculty of Business and Economics, The University of Hong Kong, China;
    \textsuperscript{4}College of Engineering, University of California, Berkeley, CA, USA
    \\

    \footnotesize \texttt{
    yjh22@mails.tsinghua.edu.cn, \{qiyongzhi1, zhangjianshen, gengdongyang, \\ chengzhengyu8, huhao\}@jd.com, qiw@tsinghua.edu.cn, maxshen@berkeley.edu
    } 
\end{center}
}
\begingroup
    \renewcommand\thefootnote{} 
    \footnotetext{\textsuperscript{*}Corresponding authors: Yongzhi Qi (qiyongzhi1@jd.com) and Wei Qi (qiw@tsinghua.edu.cn).}
\endgroup

\ABSTRACT{

Supply chain planning is the critical process of anticipating future demand and coordinating operational activities across the logistics network. However, within the context of contemporary e-commerce, traditional planning paradigms, typically characterized by fragmented processes and static optimization, prove inadequate in addressing dynamic demand, organizational silos, and the complexity of multi-stage coordination.
To address these challenges, this study proposes a fundamental rethinking of supply chain planning, redefining it not merely as a computational task, but as an interactive, integrated, and automated cognitive process. This new paradigm emphasizes the organic unification of human strategic intent with adaptive execution, shifting the focus from rigid control to continuous, intelligent orchestration. To operationalize this conceptual shift, we introduce a Generative AI-powered agentic framework. Functioning as an intelligent cognitive interface, this framework bridges the gap between unstructured business contexts and structured analytical workflows, enabling the system to comprehend complex semantics and coordinate decisions across organizational boundaries. We demonstrate the empirical validity of this approach within JD.com's large-scale operations. The deployment confirms the efficacy of this cognitive paradigm, yielding an approximate 22\% improvement in planning accuracy and a 2\% increase in in-stock rates, thereby validating the transformation of planning into an adaptive, knowledge-driven capability.

}%


\maketitle

%


\section{Introduction}

The rapid evolution of supply chain systems and the exponential growth of e-commerce impose unprecedented demands for efficiency, accuracy, and responsiveness. Modern retail operations must manage extensive product assortments, facilitate high-velocity inventory turnover, and satisfy customer expectations for expedited delivery. For instance, JD.com, a leading Chinese e-commerce platform, manages a supply chain network encompassing over 10 million self-operated Stock Keeping Units (SKUs) distributed across thousands of warehouses and distribution centers nationwide. With daily order volumes routinely exceeding tens of millions and a user base of 600 million annual active consumers, the platform demands highly coordinated inventory and logistics operations. This extensive ecosystem spans multiple interconnected stages of the value chain, ranging from supplier collaboration and assortment planning to dynamic pricing, inventory allocation, order fulfillment, and last-mile delivery. Each stage relies on precise, data-driven insights and synchronized decision-making to balance cost efficiency with service quality, thereby underscoring the immense operational complexity of modern large-scale supply chains.



\begin{figure*}[htbp]
\centerline{\includegraphics[width=0.96\textwidth]{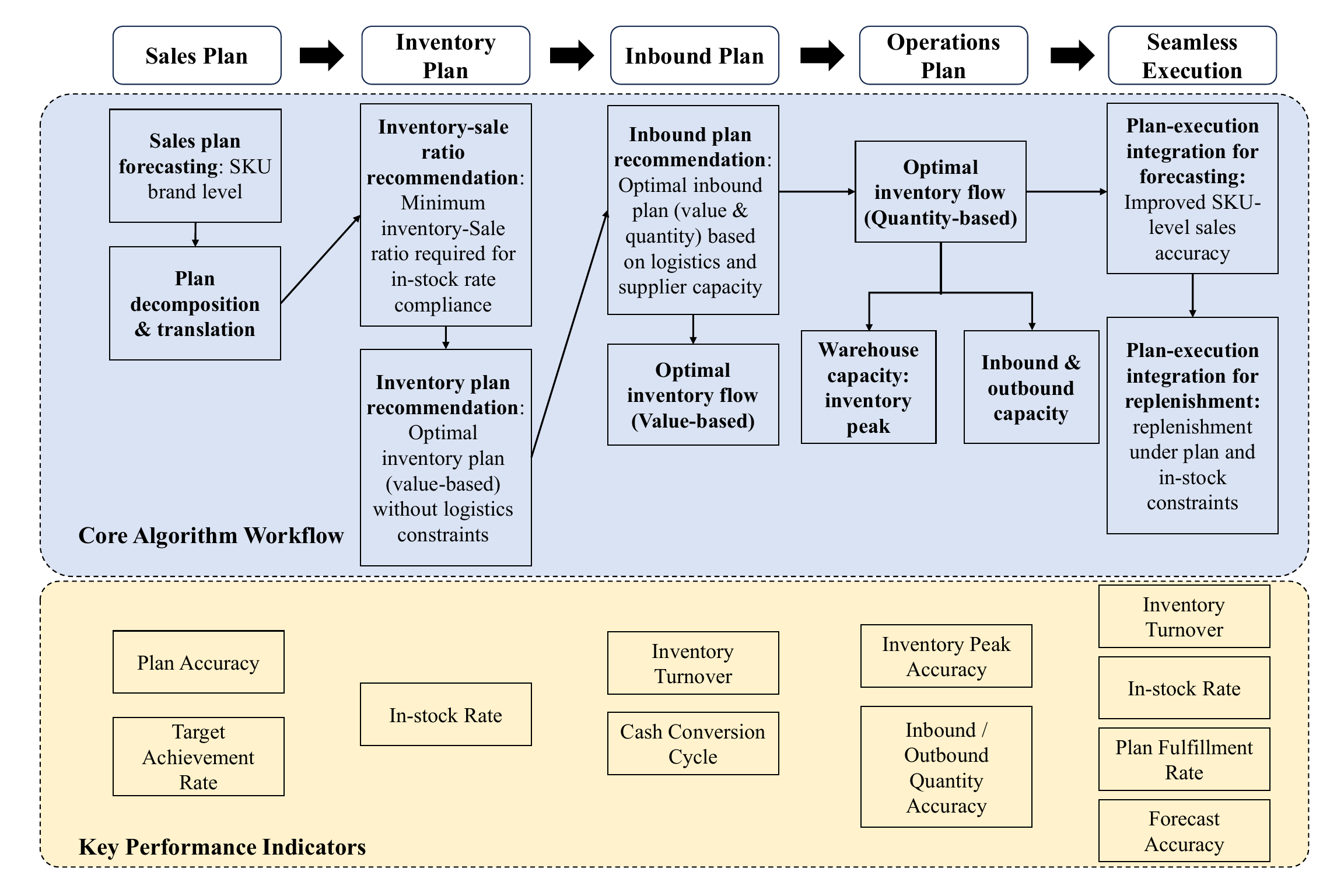}}
\caption{Workflow and Interconnections of Supply Chain Planning.}
\label{Plan_workflow}
\end{figure*}

At the core of managing this immense operational complexity lies supply chain planning. Broadly defined, this function encompasses the anticipation, organization, and coordination of future activities across the logistics network. Rather than a singular procedure, planning constitutes an integrative process spanning multiple functional departments, requiring systematic forecasting, strategic alignment, and continuous adjustment. Specifically within JD.com's operational framework, planning is categorized into four primary domains: sales, inventory, inbound logistics, and operations planning. As illustrated in Figure \ref{Plan_workflow}, each category represents a distinct decision-making layer characterized by specific stakeholders, workflows, and Key Performance Indicators (KPIs), while maintaining high interdependence with the broader system.

Effective planning is fundamental to supply chain management (SCM) as it directly influences both operational efficiency and customer service performance. Prior research indicates that well-coordinated planning processes reduce uncertainty, lower costs, and enhance responsiveness in dynamic environments \parencite{chopra2019supply,stadtler2015supply}. Specifically, planning is the primary mechanism for translating strategic objectives into executable actions across sales, inventory, procurement, and operations. Conversely, a lack of systematic planning often results in excess inventory, stockouts, and misaligned capacity utilization, all of which erode profitability and service reliability. Given the prevalence of demand volatility and lead-time variability in large-scale e-commerce supply chains, planning becomes even more critical; it provides the integrating framework necessary to synchronize decisions across multiple stakeholders and time horizons, thereby ensuring both cost efficiency and service continuity.

The contemporary landscape of supply chain planning is defined by several structural impediments:

\textbf{1. Misalignment of temporal granularity across organizational levels.}
At the strategic level, senior management establishes long-term targets, such as quarterly or annual revenue goals. In contrast, operational units require much finer-grained decisions, such as the weekly determination of product promotion priorities. The downstream challenge lies in decomposing high-level, aggregate plans into actionable, fine-grained tasks meaningful for execution. Conversely, the upstream challenge involves effectively capturing and integrating feedback from operational outcomes to inform and revise strategic plans. Traditional planning processes typically rely on repeated negotiations and manual adjustments across organizational levels, resulting in inefficiencies and delayed decision-making.


\textbf{2. Rigidity of sequential coordination across functional silos.}
Effective planning requires the synchronization of diverse functional departments, including the Economic Analysis, Procurement Control, and Operations units, which are each driven by distinct incentives and constraints. However, traditional planning relies on linear, predefined workflows: top-down financial targets must be sequentially validated and disaggregated by category management and inventory control teams. This rigid dependency structure creates significant process latency, as a bottleneck in one stage delays the entire cycle. Consequently, the reliance on pre-established, manual coordination paths renders the organization unable to swiftly adapt to dynamic market shifts, transforming necessary collaboration into a source of operational inflexibility.

\textbf{3. High labor intensity due to repetitive tasks.}
Weekly sales and inventory planning often compel analysts to manually extract, consolidate, and report data from disparate systems, calculate key metrics, or execute established algorithms. The majority of these tasks rarely involve the development of new models; instead, they rely on the repetitive application of existing tools and processes. This heavy reliance on manual effort drains significant human resources and impedes the ability of analysts to focus on higher-level analysis or strategic decision-making.

\textbf{4. Conflict between structured system requirements and flexible business needs.}
Supply chain systems mandate highly structured data formats and processes. While optimized for algorithmic calculation, this rigidity often hampers responsiveness to dynamic business requirements. For example, during major promotional periods or holidays, business teams frequently require rapid adjustment of product assortments, pricing, or allocation strategies. Furthermore, business personnel often rely on tacit experiential judgment that is difficult to communicate to analytics teams, whereas analysts must provide rigorously interpretable results to support decision-making. This information asymmetry leads to inconsistent conclusions and forces multiple rounds of iterative communication to align strategies across functional departments.


Collectively, these defining features underscore the limitations of traditional supply chain planning and create an imperative for approaches that are more adaptive, automated, and capable of integrating cross-functional insights. To address these critical challenges, recent advancements in Generative AI (GenAI) enable the transformation of supply chain planning from a rigid, system-driven process into a highly adaptive and interactive decision environment.

Generative AI synthesizes human-like text and performs complex reasoning by assimilating vast linguistic knowledge from extensive data. Such broad, cross-domain pretraining equips the technology with generalized knowledge, essential for capturing complex dependencies across diverse business functions. Furthermore, the inherent flexibility of GenAI supports rapid adaptation through dynamic interaction; such adaptive capacity facilitates sophisticated reasoning across both structured data and narrative information. These core competencies, namely contextual reasoning, comprehensive knowledge integration, and flexible adaptation, align perfectly with the requirements of supply chain planning. Specifically, the planning process demands the coordination of functional units, the interpretation of heterogeneous data streams, and the generation of actionable strategies amidst dynamic business conditions. Consequently, GenAI-powered agents establish a powerful new paradigm for automating and optimizing planning workflows within complex e-commerce operations.

Witnessing the transformative competencies of Generative AI, most notably in semantic reasoning and autonomous orchestration, compels us to fundamentally reconceptualize the definition and scope of supply chain planning.
Through a comprehensive review of historical evolution and an analysis of critical internal and external drivers, we articulate a vision where planning evolves from a static optimization tool into a generative organizational capability. We argue that under this new technological paradigm, next-generation supply chain planning functions as an intelligent organizational interface that integrates semantic understanding, orchestrates cross-functional collaboration, and adapts dynamically to uncertainty. To realize this vision, we introduce a GenAI-powered agentic framework. This architecture serves as a practical implementation of the proposed paradigm, demonstrating how intelligent agents effectively bridge the gap between strategic intent and operational execution within a complex, real-world e-commerce environment.


We collaborate with JD.com, a leading Chinese online retailer, to investigate the planning complexities inherent in a large-scale e-commerce supply chain. We systematically model the enterprise’s planning processes to capture the intricate interactions among diverse business categories and functional departments. Furthermore, we extract and formalize key workflows to precisely identify critical decision points, latent data dependencies, and persistent operational challenges. Through this comprehensive modeling, we uncover significant inefficiencies and coordination difficulties, establishing a problem-driven foundation for designing solutions that directly address real-world planning requirements.



We summarize our main contributions as follows:

\begin{enumerate}
    \item \textbf{A reconceptualization of supply chain planning.} 
    Synthesizing the historical evolution of planning with contemporary operational complexities, this study advances a fundamental reconceptualization of supply chain planning. We redefine supply chain planning as \textit{an interactive, integrated, and automated generative process that systematically produces vertically coherent, horizontally synergistic, and dynamically adaptive plans across organizational and temporal layers}. This definition shifts the paradigm from static, rule-based optimization to cognitive orchestration, positioning planning as an intelligent organizational interface. By explicitly addressing the need for vertical coherence and horizontal synergy, this theoretical framework provides the essential foundation for leveraging Generative AI not merely as a computational tool, but as a core enabler of organizational resilience and agility.
    \item \textbf{A Generative AI-powered agentic architecture for cognitive orchestration.}
    To materialize our reconceptualization of supply chain planning, we develop a generalizable GenAI-powered agentic framework that operationalizes the proposed cognitive paradigm.
    This architecture bridges the critical gap between strategic intent and operational action by integrating specialized cognitive mechanisms, specifically semantic interpretation, autonomous task decomposition, and iterative reasoning. Unlike general-purpose language models, this framework aligns agent reasoning with domain-specific constraints and heterogeneous data sources, offering a systematic methodology for embedding Generative AI into core organizational workflows to achieve closed-loop decision automation.
    
    \item \textbf{Empirical validation in a large-scale e-commerce environment. }
    The implementation of the proposed framework proves to be a success. Deployed and validated within JD.com's massive supply chain network, the system demonstrates its practicality and robustness in a high-velocity, real-world setting. Empirical results confirm significant operational improvements: the system reduces weekly data processing time by approximately 40\%, increases planning accuracy by 22\%, and improves stock fulfillment rates by 2–3\%. These findings provide tangible evidence that GenAI-powered agents can transcend conceptual potential to deliver measurable business value and drive the transformation toward intelligent supply chain orchestration.
    
    
\end{enumerate}


\section{
Retrospect and Redefinition
}

Contemporary supply chain planning necessitates a renewed conceptual foundation. We commence by reviewing the historical evolution of the planning concept, tracing its trajectory through key stages shaped by technological advances and business shifts. Subsequently, we examine the converging internal and external forces driving contemporary transformations—ranging from the imperative for process redesign and market volatility to the breakthroughs enabled by Generative AI. The discussion culminates in presenting our thoughts on redefining supply chain planning in the era of Generative AI.

\subsection{Evolution of the Planning Concept}
In the early context of management science, planning is primarily regarded as a deterministic problem-solving tool. \textcite{orlicky1974material} introduces the Material Requirements Planning (MRP) system, which formalizes planning as optimal allocation tasks under resource constraints and provides enterprises with precise methods for production and inventory control. Subsequently, \textcite{hax1973hierarchical} and \textcite{bitran1993hierarchical} further develop the theory of hierarchical planning, enabling structured implementation from the strategic level down to operational execution.
The logic of planning during this stage relies heavily on predictability: the external environment is assumed stable, information flows unidirectionally, and the system’s core objective is to achieve control and optimality via algorithmic computation. This emphasis reinforces the perception of planning as a precise and reliable management tool.

However, with the emergence of the supply chain concept, the boundaries of planning are fundamentally expanded. \textcite{lee1992managing,lee1993material} highlight that the essence of a supply chain lies in coordinating information and material flows across organizational boundaries, rather than optimizing solely within a single enterprise. Subsequently, \textcite{simchi1999designing} and \textcite{stadtler2015supply} introduce the Advanced Planning and Scheduling (APS) systems, defining planning as a mechanism for horizontal and vertical coordination across hierarchical levels and organizational boundaries. During this stage, planning evolves from an internal control system into a networked collaborative system, where its core value shifts from mere efficiency to coordination and end-to-end visibility across the supply chain.

Entering the 21st century, supply chains have faced a dramatic surge in complexity and volatility. In response, \textcite{lee2004triple} introduced the influential ``AAA'' framework, defined by Agility, Adaptability, and Alignment, which serves as a pivotal milestone in the evolution of modern planning.
The AAA framework implies a fundamental shift: planning is no longer mere resource allocation within fixed cycles, but rather a mechanism for organizational response to uncertainty. \textcite{ivanov2020viability} further introduces the concept of a ``viable supply chain,'' emphasizing that planning systems must possess dynamic reconfiguration capabilities, enabling the generation of new configurations and plans as the environment changes. This indicates that the function of planning evolves from ``formulating an optimal plan'' to ``continuously generating feasible plans.''

Where early planning represents an attempt to control complexity, modern planning is redefined as a cognitive interface leveraging organizational intelligence. In the context of modern e-commerce and smart supply chains, planning systems must transcend simple data processing to comprehend semantics, identify user intentions, generate hypotheses, and interact naturally with human operators. Planning thus evolves from static optimization algorithms to an agentic system capable of natural language understanding, continuous feedback assimilation, and proactive solution generation. This transformation also endows planning with a generative logic: the process no longer assumes deterministic inputs but dynamically generates the decision space through continuous learning and semantic modeling.

Traditionally, supply chain planning is carried out through a combination of rule-based systems, human expertise, and optimization models such as Mixed Integer Programming (MIP) or heuristics.
Rule-based systems encode domain knowledge into fixed decision rules, enabling automated plan generation under stable conditions. Optimization models, on the other hand, aim to compute mathematically optimal plans given a well-defined objective function and a set of constraints.
Although these approaches are effective in well-structured, static environments, they face several limitations in modern e-commerce scenarios:
(i) \textbf{Data heterogeneity and incompleteness.} Input data originates from various sources, including Enterprise Resource Planning (ERP), Warehouse Management System (WMS), Transportation Management System (TMS), and third-party logistics systems, and often suffers from delays, noise, or missing values, complicating the analysis and decision-making process. 
(ii) \textbf{High uncertainty.}
Demand patterns are volatile due to factors such as market trends, seasonality, and external disruptions, making it difficult to forecast accurately and plan efficiently.
(iii) \textbf{Execution inefficiency.}
Once a plan is generated, adjustments to accommodate new constraints or changing business priorities often require restarting the planning process from scratch, resulting in significant delays and inefficiencies. 
(iv) \textbf{Adaptation latency.}
Traditional planning systems often involve time-consuming recalculations or manual interventions when external conditions change, leading to slower response times and reduced flexibility in dynamic environments.

\subsection{Internal Drivers}
Having traced the historical evolution of planning paradigms, we next examine the fundamental shifts in firms' internal structures and processes that primarily drive the transformation of modern supply chain planning.
This transformation not only explains the past sufficiency of traditional planning systems in stable environments but also reveals why next-generation planning must embody new forms of cognition and interaction. As \textcite{lee2021new} argues in his evolved AAA framework, the contemporary business environment demands a fundamental upgrade of agility, adaptability, and alignment, entailing an inward-to-outward reconstruction of organizational capabilities. This reconstruction involves dynamically modeling uncertainty, dismantling rigid Sales and Operations Planning (S\&OP) cycles, and systematically integrating managerial judgment into planning processes. Similarly, the concept of the Autonomous Supply Chain (ASC), proposed by \textcite{xu2024towards}, reflects this paradigm shift, emphasizing the need for structural autonomy in supply chain planning—an evolution that is not merely technological but also conceptual. From another perspective, \textcite{tang2019strategic} demonstrate that the essence of logistics transformation in the Industry 4.0 era lies in a strategic reorientation from ``risk avoidance'' to ``failure tolerance and rapid recovery,'' which imposes similar demands on the design of modern planning systems.


A key challenge for modern supply chain planning lies in \textbf{overcoming entrenched information silos and achieving effective data integration}. Key operational divisions, including sales teams, inventory planners, procurement managers, and production units, typically operate their data systems independently. This results in information flows that are mostly unidirectional and confined within departmental boundaries, rendering localized optimization sufficient for traditional planning systems. However, in today's large-scale e-commerce enterprises, such as Amazon, Walmart, and JD.com, operations span multiple regions and product categories. This proliferation yields highly heterogeneous data sources, ranging from structured order data and semi-structured logistics reports to unstructured customer feedback, thereby significantly compounding the complexity of planning and coordination \parencite{shen2025jd, hu2024supercharged, zhang2025unstructured}.


Empirical studies underscore the importance of integration in enhancing supply chain responsiveness and resilience. \textcite{tarigan2021impact} demonstrate that internal integration positively influences sustainable competitive advantage by strengthening supply chain partnerships, agility, and resilience. 
\textcite{raj2025advancing} introduce the notion of Supply Chain Hyper-Agility, emphasizing that internal information integration becomes particularly critical in enabling rapid response during extreme disruptions. A practical illustration stems from IKEA's supply chain planning transformation: as analyzed by \textcite{jonsson2013centralised}, the company consolidates planning responsibilities, previously distributed across 120 regional departments, into a centralized team of around 30 planners. This action significantly improves forecasting accuracy and inventory transparency, showcasing the tangible benefits of unified planning systems.

%
Moreover, \textbf{the seamless integration of data and decision execution} becomes a necessary condition for modern supply chain planning. Early planning systems emphasize a linear ``plan–execute–feedback'' process, where delays between decision-making and execution have limited impact in stable environments. However, in today's volatile markets, real-time responsiveness is essential. For instance, during JD.com's ``Double 11'' shopping festival, the total transaction value reaches approximately RMB 1.44 trillion \parencite{syntun2024double11}, making real-time inventory and transportation adjustments critical to ensuring fulfillment efficiency.


Supply chain resilience demands a fundamental shift in planning objectives. \textcite{simchi2014superstorms} posits that planning systems must evolve beyond pure cost minimization to incorporate structural flexibility and stress-testing capabilities. They argue that robust planning requires the ability to simulate disruption scenarios and prepare mitigation strategies ex-ante, rather than relying solely on reactive adjustments. This theoretical imperative is vividly illustrated by the COVID-19 pandemic, which exposes the inadequacy of static planning logic. In the case of JD.com, extreme demand fluctuations and network interruptions force a rapid transition from rigid schedules to dynamic logistics replanning, requiring real-time ecosystem collaboration to stabilize supply \parencite{shen2023strengthening}. Complementing this, \textcite{aljohani2023predictive} highlights that contemporary planning shifts from post-event analysis to proactive scenario generation, enabled by predictive analytics. Collectively, these factors transform supply chain planning from a deterministic optimization task into an intelligent organizational interface designed to navigate uncertainty.


\textbf{Cross-functional collaboration with competing objectives} renders the coordination value of planning more salient than ever. In traditional firms, where internal functions are relatively singular, planning is primarily concerned with the local optimization of internal resources. In modern supply chain environments, however, departments differ markedly in their objectives, Key Performance Indicators (KPIs), and operational constraints. For example, warehouse teams emphasize inventory turnover and capacity utilization; procurement teams prioritize cost and supply stability; while sales teams focus on order fulfillment rates and category strategies. In complex omnichannel retail settings, empirical evidence shows that deep coordination across organizational structures, business processes, and information systems is a critical driver of supply chain resilience and overall performance \parencite{zhang2022green}.

As Generative AI (GenAI), machine learning (ML), and predictive analytics become increasingly embedded in supply chain operations, the role boundaries of internal positions are being fundamentally redefined. Many tasks that previously rely on rules or experience shift toward data-driven decision support, cross-functional coordination, and intelligent oversight \parencite{koneti2024human}. This evolution implies that modern planning systems are no longer mere efficiency-enhancing tools; they function as coordination mechanisms that align cognition, information, and incentives across departments. Planning becomes a central conduit for internal information flow, semantic understanding, and intent alignment, enabling organizations to continuously generate more coherent and resilient decision outcomes amid goal conflicts and resource constraints.


Finally, \textbf{the digitalization of tacit knowledge and experiential know-how} becomes a critical foundation for intelligent supply chain planning. Traditional planning relies heavily on experienced managers and limited historical data; however, as business complexity increases, operational heuristics and experiential rules are increasingly dispersed across individuals and large volumes of unstructured documents, rendering systematic codification difficult. Modern planning systems enhance organizational decision intelligence by structuring tacit knowledge, applying semantic modeling, and leveraging generative capabilities, thereby improving supply chain resilience and adaptability \parencite{tarigan2021impact}.

\textcite{herden2020explaining} emphasizes that the competitive advantage derived from analytical capabilities does not stem from technology alone but is rooted in the systematic construction of knowledge integration and application mechanisms. The strategic incorporation of external knowledge sources is essential for accelerating the digitalization of experience. The digital transformation trend receives accelerated validation and external impetus under the supply chain disruptions caused by COVID-19. Firms must establish robust absorptive capacity for external knowledge and ensure its coordination with internal knowledge integration mechanisms \parencite{ngo2023digital}. Under the framework of Supply Chain 4.0, the digitalization of experiential knowledge is a socio-technical dual-track system, requiring simultaneous advances in technological intelligence and organizational process redesign to shift planning from being ``experience-driven'' to ``knowledge-driven'' \parencite{garcia2023creating}.

Overall, these internal factors reveal a fundamental tension: traditional planning is ``sufficient'' largely because internal environments are simple, stable, and predictable. Moving forward, supply chain planning must be upgraded because internal structures are more complex, information silos remain pervasive, and the need for data–decision–execution integration is intensifying. Planning is no longer a mere optimization tool; it functions as the cognitive interface of the organization, capable of dynamically generating plans, interpreting semantic signals, coordinating across functions, and enabling continuous adaptation and decision innovation.

\subsection{External Forces}
Complementing the internal reconstruction of organizational capabilities, the transformation of supply chain planning is equally propelled by a distinct set of external forces, beginning with the evolving regulatory landscape.
\textbf{Government policies and regulatory frameworks} serve as key external drivers for modern supply chain planning. These requirements cover sustainability, environmental protection, and cross-border trade, directly shaping the constraints and data demands of planning systems. For example, the European Union Emissions Trading System (EU ETS) requires companies to control carbon emissions in production, transportation, and inventory processes, compelling supply chain plans to incorporate environmental metrics and generate traceable emissions schedules \parencite{EU_ETS}. Similarly, China's Carbon Peak and Carbon Neutrality Action Plan encourages manufacturers and logistics companies to consider energy consumption and carbon footprint in production scheduling, transportation planning, and inventory management \parencite{China_Carbon}. Frequent shifts in cross-border trade policies, exemplified by the tariff adjustments during the US–China trade tensions, compel multinational companies to dynamically recalibrate their production plans, inventory allocation, and transportation routes.
Additionally, policy-driven support provides external resources to promote upgrades in supply chain planning systems. For instance, policies on smart logistics and Industry 4.0 encourage firms to implement APS systems, intelligent warehousing, and predictive analytics, while digital transformation subsidies facilitate the rapid deployment of intelligent planning solutions. Carbon taxes and environmental compliance regulations further require firms to balance cost and environmental objectives, ensuring alignment between internal decision-making and external policy requirements \parencite{WorldBank_CarbonPricing}. Overall, government policies not only set the constraints that supply chain planning must satisfy but also provide resources and incentives, transforming planning from a traditional static optimization tool into a dynamic, visible, and generative intelligent system.

Another external force reshaping supply chain planning arises from \textbf{broader industry and ecosystem dynamics}. Competitive industry structures, platform-based business ecosystems, and cross-enterprise alliances increasingly push firms to adopt advanced planning methods. The core insight is that industry-level ecosystems provide shared resources, technological infrastructures, and proven practice templates, thereby reducing the marginal cost for firms to upgrade their planning capabilities. Industry demand for economic performance drives firms toward more responsive and analytically sophisticated planning. A well-known illustration is Zara’s highly responsive supply chain, where tight integration of design, production, and distribution demonstrates how superior planning translates directly into faster cycle times, lower inventory risk, and higher profitability \parencite{aftab2018super,scmglobe2024}. Similarly, Intel builds an end-to-end analytics system that spans product architecture design and supply chain planning, reducing engineering and manufacturing costs while improving customer responsiveness and resource utilization, ultimately generating roughly \$25 billion in economic impact \parencite{heiney2021intel}. These cases show that in fast-paced sectors, advanced planning is not optional but a determinant of competitive advantage.

Multi-party collaboration requirements, particularly those involving upstream suppliers, further elevate the need for modern planning systems. The sustainability collaboration between Unilever and Walmart demonstrates how supplier–retailer alliances increasingly rely on shared data, coordinated planning of sourcing and logistics, and collaborative decision-making to achieve efficiency and compliance goals \parencite{walmart2018}. Such cases highlight how planning must extend beyond organizational boundaries and support cross-enterprise orchestration.
Industry-wide digital transformation trends accelerate the adoption of new planning paradigms. Platform providers such as Amazon show how machine-learning-based lead-time insights can materially improve supply planning accuracy, illustrating how cloud ecosystems are shaping planning best practices at scale \parencite{aws2023}. These examples indicate that industry trendsetters create both competitive pressure and practical roadmaps for others to follow.
Collectively, these industry and ecosystem forces create a powerful external impetus for redefining supply chain planning—from an internal optimization task to a network-embedded, data-driven, collaborative, and continuously evolving capability.


Significantly,\textbf{ technological disruption, most notably the emergence of Generative AI}, acts as a critical external force that is fundamentally reshaping modern supply chain planning. Exemplified by foundation models like GPT-3, GPT-4 \parencite{brown2020language, achiam2023gpt}, Generative AI transcends traditional analytical limits by introducing advanced semantic reasoning and autonomous task orchestration. Unlike static calculation methods, these capabilities enable the system to decompose complex planning problems and transform planning into a dynamic reasoning process. This shift allows organizations to conduct rigorous ``what-if'' analyses under uncertainty, significantly improving both forecasting accuracy and operational efficiency \parencite{prakash2025generative}.

To operationalize these capabilities, a new architectural paradigm emerges: Generative Agents. These agents represent an advanced class of artificial intelligence that combines the linguistic strengths of foundation models with task-specific execution capabilities \parencite{zhu2024large}. Structurally, they consist of a core reasoning engine integrated with specialized functional components, allowing them to interact with both structured and unstructured data. This architecture enables agents to operate autonomously in dynamic environments, bridging the gap between abstract reasoning and real-time decision support.

In the domain of supply chain management, empirical studies demonstrate the transformative potential of these generative capabilities across diverse operational contexts. One prominent application involves deploying Generative AI models as ``supply chain interpreters,'' enabling business users to drive complex optimization systems through natural language queries. As demonstrated by \textcite{simchi2025large}, this approach significantly improves decision efficiency and democratizes access to advanced planning tools. Similarly, studies on the Microsoft Cloud supply chain highlight how generative models facilitate interpretability in data-driven decision-making \parencite{srivastava2024exploring}. Regarding human-AI collaboration, \textcite{venkatachalam2025integrating} illustrate how agents augment human planners by providing natural-language summarization and conversational KPI adjustments.

Furthermore, recent literature highlights the efficacy of agents in autonomous simulation and proactive resilience. \textcite{quan2024invagent} introduce InvAgent, a framework that utilizes zero-shot learning to construct agent-based simulations, demonstrating superior adaptability in dynamic inventory management. Similarly, \textcite{jannelli2024agentic} employ agents to execute core operational activities, including ordering and fulfillment, thereby mitigating the bullwhip effect. From a resilience perspective, generative models facilitate proactive risk management by simulating disruption events and formulating adaptive mitigation strategies \parencite{riad2024enhancing}. Broader reviews confirm that this trend extends beyond forecasting to dynamic network reconfiguration \parencite{teixeira2025intelligent, daios2025ai}, while \textcite{li2023large} emphasize the critical need for robust infrastructure and foundation models to support the broader ecosystem of generative applications.

Collectively, the confluence of evolving regulatory landscapes, competitive ecosystem dynamics, and the transformative disruption of Generative AI forges the critical external forces reshaping the discipline. Stringent regulatory mandates and hyper-competitive market pressures impose uncompromising demands for agility and compliance; simultaneously, Generative AI shatters historical barriers by equipping organizations with unprecedented cognitive capabilities. This synergy of pressure and capability renders static optimization obsolete, creating an undeniable imperative to rethink supply chain planning as a dynamic, intelligent, and generative organizational capability.

\subsection{A New Definition}

Having traced the historical evolution of supply chain planning and analyzed the converging internal and external forces driving its transformation, we synthesize these observations and offer the following perspective on its conceptual renewal.

We propose a renewed conceptualization of supply chain planning as \textit{an interactive, integrated, and automated generative process that systematically produces vertically coherent, horizontally synergistic, and dynamically adaptive plans across organizational and temporal layers}. Rather than treating planning as a linear or siloed set of tasks, this perspective regards it as a cohesive system that continually aligns decisions with evolving business objectives and operating conditions. The process organically unifies human strategic intent, semantic understanding of operational contexts, and machine-driven analytical and execution capabilities. It orchestrates demand forecasting, inventory management, procurement, production, distribution, and fulfillment to improve system-wide efficiency, responsiveness, and resilience. Enabled by continuous human–machine collaboration, real-time information integration, and adaptive optimization, this conceptualization positions modern planning as a dynamic, self-reinforcing mechanism for navigating increasingly volatile environments.

From the \textit{interactive} perspective, planning should no longer remain a unidirectional set of instructions or static spreadsheets; it must transform into a dynamic dialogue where the system interprets user requirements, provides feedback on inferred outcomes, and continuously mediates between decisions and data. From the \textit{integrative} perspective, future planning needs to dismantle traditional functional silos, aligning departments, systems, and data within a shared semantic framework to ensure consistency in information, objectives, and constraints. From the \textit{automation} perspective, planning systems must advance beyond executing predefined rules—they should be capable of self-learning, generating decision logic, and optimizing plans autonomously.

Under this new paradigm, supply chain planning evolves from a static optimization activity into a generative reasoning process. Planning shifts from ``control'' to ``co-evolution,'' establishing a continuous loop of collaboration among humans, data, and intelligent systems. This approach enables dynamic generation of decisions, adaptation to environmental changes, and ongoing optimization. 
Ultimately, such a planning system will not only empower organizations to navigate deep uncertainty and complexity but will also serve as the catalyst for enduring supply chain resilience and innovation.

Building on this new notion of supply chain planning, we materialize the proposed paradigm by developing a Generative AI-powered agentic framework designed to function as an intelligent organizational interface. To demonstrate the empirical efficacy of this architecture under the constraints of modern supply chain planning, we collaborate with JD.com, the largest e-retailer in China by revenue. The remainder of this paper is organized as follows: Section 3 details the technical architecture and autonomous orchestration mechanisms of the proposed framework, while Section 4 presents the real-world deployment and performance validation within JD.com’s operational scenario.



\section{A Proposed Agentic Framework}

\subsection{Recalibration of the Supply Chain Planning Process}

Building upon the redefinition of supply chain planning, it is imperative to recalibrate the operational workflow to align with this conceptual shift. While the proposed paradigm emphasizes an interactive and integrated nature, this recalibration does not imply collapsing the workflow into a monolithic or undifferentiated black box. Conversely, contemporary supply chain operations, characterized by data-rich and high-velocity environments, require the rigorous decomposition of planning activities into discrete stages to effectively facilitate end-to-end coordination. Crucially, these stages operate within a unified cognitive environment, facilitating continuous feedback, adaptation, and symbiotic collaboration between human decision-makers and intelligent agents.

Within the context of e-commerce supply chain management, we reconfigure planning as a closed-loop, multi-stage cycle that encompasses the entire lifecycle from data synthesis to execution monitoring and adaptive correction. Specifically, a holistic planning workflow comprises the following integrated stages:

\begin{enumerate}
    \item \textbf{Autonomous Data Acquisition}: Leveraging the agent's capacity to navigate the massive dimensionality of e-commerce operations. Unlike human planners who are limited to aggregated views, the agent utilizes high-throughput tool invocation to execute rapid, granular queries across millions of SKUs. 
    By autonomously synchronizing vast and heterogeneous datasets, encompassing real-time sales streams, commercial targets, and intricate PSI schedules, the system effectively resolves information complexity, ensuring that planning is driven by full-spectrum, zero-latency business signals.
    \item \textbf{Plan Generation}: Employing the agent's Chain-of-Thought (CoT) reasoning to bridge the gap between strategic intent and operational execution. Unlike traditional ``black-box'' solvers, the agent semantically interprets high-level business objectives and logically decomposes them into granular, department-specific sales and PSI targets. This ensures that the generated baseline is not merely a numerical output, but a rationally derived strategy that maintains strict alignment with organizational goals.
    
    \item \textbf{Plan Execution}: Bridging the critical gap between strategic planning and tactical execution. The agent functions as an intelligent orchestrator, employing context-aware tool invocation to translate high-level planning targets into precise, system-level directives (e.g., purchase orders and parameter updates). By embedding domain logic into this translation process, the system ensures that the top-down implementation is not only operationally viable but remains strictly aligned with the original strategic intent, preventing the dilution of business objectives during execution.

    \item \textbf{Plan Diagnosis and Early Warning}: Providing dynamic sense-making amidst operational volatility. The agent moves beyond static variance reporting to actively navigate execution uncertainty. By employing real-time causal reasoning, it addresses the critical challenge of disentangling ambiguous signals, such as autonomously distinguishing between a transient sales fluctuation and a structural market shift. This mechanism functions as a continuous vigilance loop, ensuring that the system dynamically identifies and isolates risks before they escalate into systemic failures.
    
    \item \textbf{Plan Correction}: Closing the adaptive loop through autonomous remedial action. Upon receiving diagnostic signals, the agent leverages generative simulation to synthesize and validate optimal recovery strategies (e.g., inventory rebalancing or expedited replenishment). Instead of rigid rule-based adjustments, it employs counterfactual reasoning (``what-if'' analysis) to predict the outcome of corrective measures, ensuring that the revised plan effectively recalibrates the system against market shifts and execution uncertainties.
\end{enumerate}

This structured workflow instantiates the proposed ``Intelligent Organizational Interface.'' By requiring semantic reasoning and autonomous adaptation across all stages, spanning from data synthesis to corrective action, this cycle surpasses static calculation. Such operational complexity necessitates the deployment of Generative AI-powered agents, which are uniquely equipped to navigate these reasoning demands. Consequently, this architecture enables the continuous recalibration of strategies, dynamically balancing efficiency and resilience amidst market volatility.


\begin{figure*}[htbp]
\centerline{\includegraphics[width=0.96\textwidth]{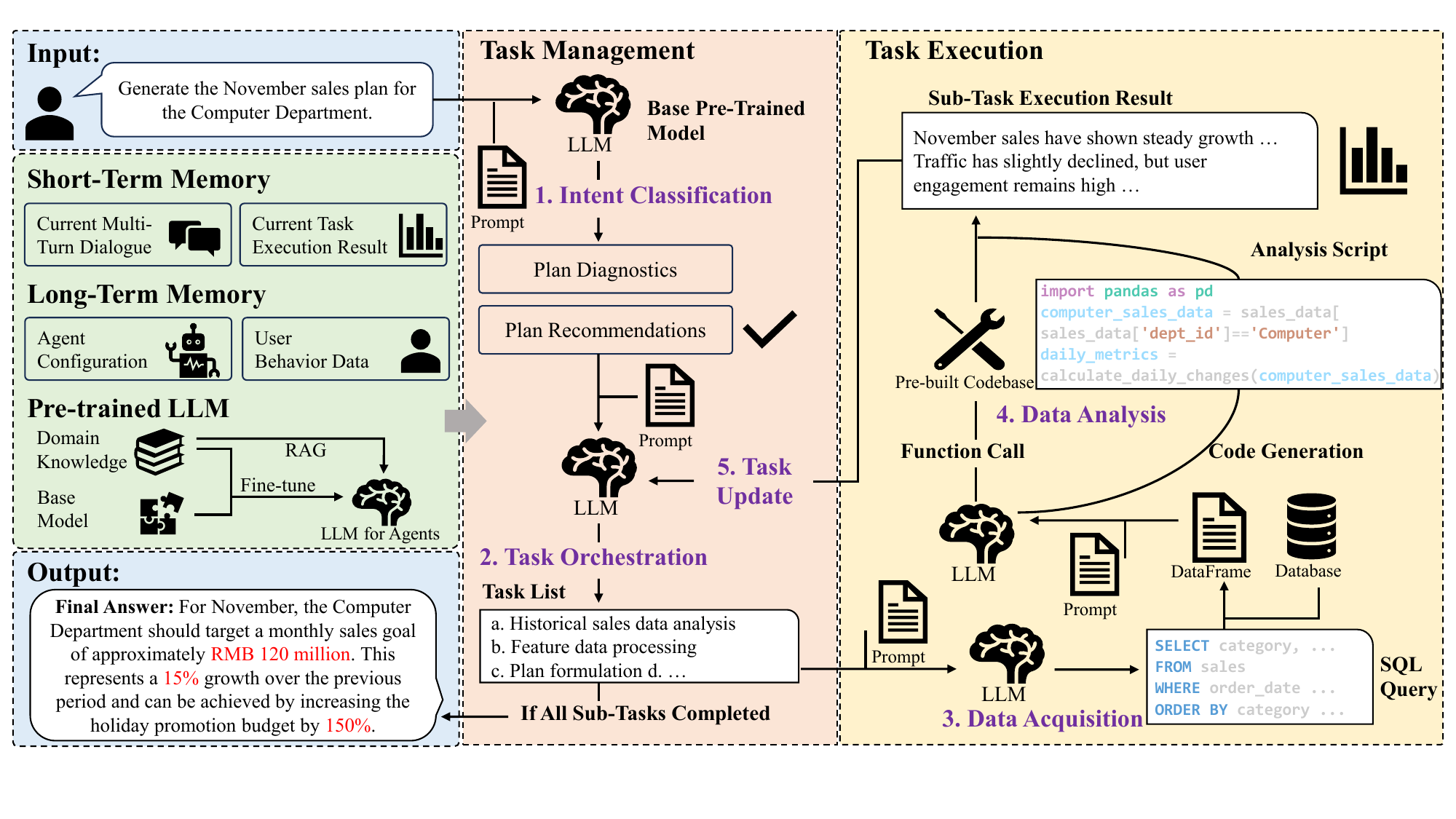}}
\caption{The proposed agentic framework.}
\label{agent_framework}
\end{figure*}

Synthesizing the essential elements required to balance human strategic intent with machine-driven execution, we propose the \textbf{Generative AI-powered Supply Chain Planning Agent Framework} (Figure \ref{agent_framework}) to operationalize this recalibrated workflow. Designed as an intelligent organizational interface, this architecture bridges the gap between business intent and technical execution by empowering domain experts to drive complex planning cycles directly through natural language. The system functions as a cognitive orchestrator, decomposing high-level directives, such as ``Generate a sales plan for November,'' into a structured, multi-agent workflow. Each workflow stage is managed by a specialized sub-agent, underpinned by domain-adapted models and task-specific prompts, ensuring that semantic reasoning is rigorously translated into precise operational actions.


The proposed architecture comprises two primary modules: \textbf{Task Management} and \textbf{Task Execution}. The \textbf{Task Management module} functions as the cognitive control unit, interpreting natural language inputs and decomposing these directives into executable sub-tasks. The critical challenges within this domain include precise semantic recognition and robust task reasoning. Complementing this orchestration, the \textbf{Task Execution module} manages the implementation of the generated task list and synthesizes the corresponding outputs. The primary technical hurdles in this phase involve the seamless integration of heterogeneous databases for data extraction and the capability to dynamically generate execution code for novel tasks undefined in the existing repository.

In the subsequent sections, we detail the functional architecture of the proposed framework. We systematically define the specific role of each sub-agent, mapping the transformation of data inputs into decision outputs at every stage of the planning lifecycle. This component-level analysis clarifies the mechanisms through which the system achieves autonomous orchestration and domain-specific adaptation.
To provide a granular view of this workflow, we present a comprehensive case study in the Appendix, which illustrates the specific inputs and outputs generated at each step and demonstrates how the agents collaborate to transform strategic intent into executable actions.

\subsection{The Foundation Model}



Before detailing the specific agent workflows, it is essential to establish the role of the Foundation Model as the unified cognitive nucleus of the entire framework. As depicted in the left panel of Figure \ref{agent_framework}, this model functions as the shared ``brain'' that powers every subsequent agent, providing the fundamental reasoning capabilities required for task decomposition and execution. To transform this general-purpose model into a specialized supply chain expert, we integrate it with critical enabling mechanisms: \textit{short-term memory} for maintaining interaction context, \textit{long-term memory} for retrieving domain knowledge, and \textit{fine-tuning} for aligning with strategic objectives. These components are not merely additive features but are the foundational prerequisites that equip the reasoning engine with the stability and depth necessary to drive the multi-agent collaboration.

\begin{figure*}[htbp]
\centerline{\includegraphics[width=0.96\textwidth]{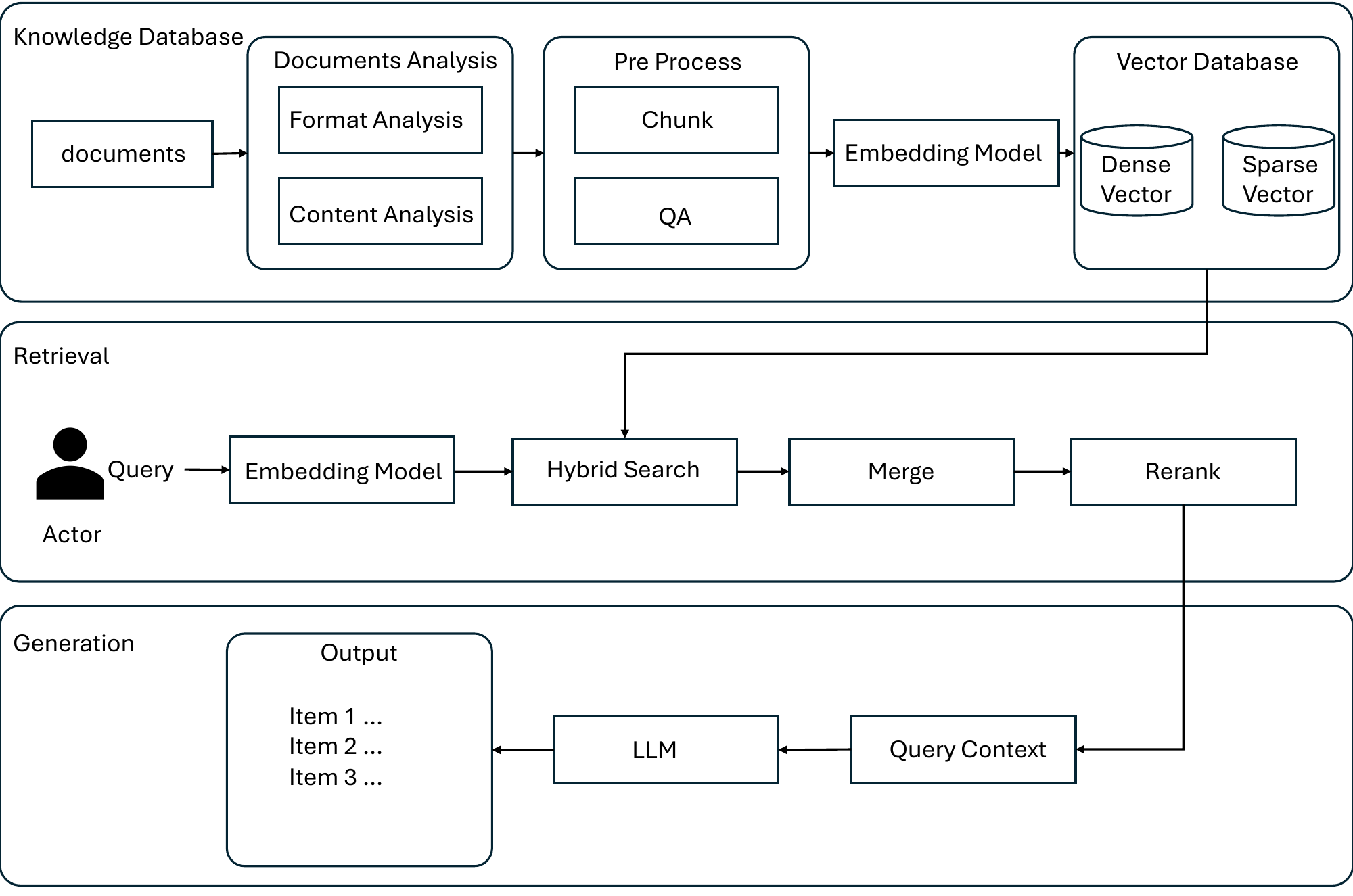}}
\caption{Domain knowledge and data generation.}
\label{database}
\end{figure*}


\textbf{Short-term memory} enables the reasoning engine to maintain awareness of the current conversational context and intermediate task states. This transient storage mechanism preserves continuity across multiple exchanges with the user or between agents, ensuring coherence and goal alignment during multi-step planning. For example, specific details regarding the current planning objective or prior task decisions are stored temporarily to guide subsequent reasoning and execution.

In contrast, \textbf{long-term memory} functions as a persistent repository for domain knowledge and accumulated interactions. This memory component stores user preferences, historical decisions, and validated results from prior sessions, enabling the system to construct a progressively richer understanding of the business context over time. Such persistence is crucial for supply chain planning scenarios, where strategic decisions are inherently sequential and deeply influenced by historical performance.



To enhance reasoning accuracy and domain alignment, the architecture integrates \textbf{retrieval-augmented generation (RAG) and fine-tuning}. Figure \ref{database} illustrates the knowledge base construction, detailing how heterogeneous documents undergo content analysis and chunking to populate a hybrid vector database. By employing dense and sparse embedding strategies followed by reranking mechanisms, this pipeline connects the reasoning engine to real-time enterprise contexts. Complementing this, fine-tuning on specialized corpora aligns the model's decision logic with supply chain protocols. Collectively, these strategies anchor general reasoning in factual domain intelligence, establishing a robust foundation for the subsequent multi-agent orchestration.



\subsection{Query Enhancement}\label{query_enhance_section}
\begin{figure*}[htbp]
\centerline{\includegraphics[width=0.96\textwidth]{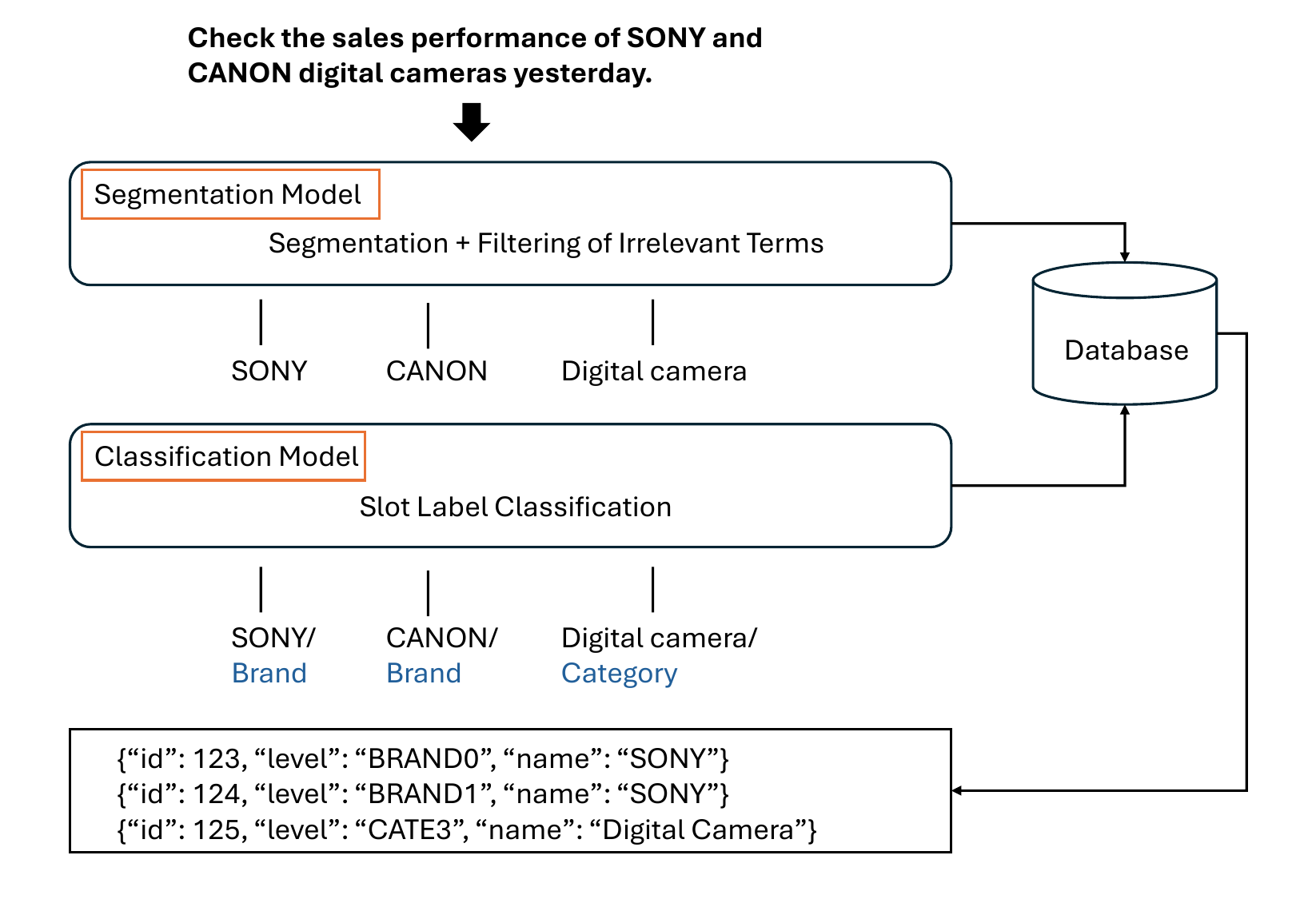}}
\caption{Query enhancement.}
\label{query_enhance}
\end{figure*}

Leveraging the semantic processing capabilities of the underlying Foundation Model, the workflow initiates with a Query Enhancement Module. Designed to mitigate the ambiguity inherent in natural language, this module preprocesses raw user inputs, transforming them into structured, machine-parsable representations suitable for downstream orchestration. The primary objective is to distill critical planning parameters from unstructured queries, thereby maximizing both the interpretability of the user's intent and the execution accuracy of subsequent agents. Figure \ref{query_enhance} presents a concrete case illustration, demonstrating the step-by-step transformation of a raw natural language query into a finalized, structured directive.

This module can be regarded as a typical Natural Language Processing (NLP) task and primarily consists of two sub-models:

\textbf{1. Tokenization and Filtering Model:}
   This sub-model segments the user’s original input into meaningful tokens and filters out noise or irrelevant words (e.g., stop words and redundant conjunctions) using predefined dictionaries and semantic rules. Through this process, the system preserves only the semantically valuable entities and action-related keywords, providing a cleaner and more informative representation for subsequent intent recognition and task decomposition.

\textbf{2. Slot Classification Model: }
   After tokenization and filtering, the system employs a classification model to map each token to a specific slot, identifying its semantic role within the query, such as time, product brand, or category. This step bridges the gap between text-level semantic parsing and structured semantic representation, enabling a more precise understanding of user intent.

By integrating these two sub-models, the Query Enhancement Module effectively strengthens the system’s ability to comprehend complex natural language queries and provides high-quality input features for the downstream intent recognition, task decomposition, and execution modules.

\subsection{Task Management Agents}

Upon converting the raw user input into a structured representation, the workflow advances to the Task Management module, which serves as the operational bridge between abstract intent and concrete execution. Driven by specialized agents utilizing fine-tuned Large Language Models and context-specific prompts, this component executes a collaborative two-stage process. First, the Intention Classification agent interprets the structured query to disambiguate the underlying business goal, ensuring precise alignment with the domain context even under ambiguity. Subsequently, the Task Orchestration agent decomposes this identified intent into a coherent, dependency-aware workflow, translating high-level objectives into a sequence of actionable operational tasks. By grounding these agents in domain-specific fine-tuning and tailored prompting strategies, the framework establishes a flexible reasoning mechanism capable of navigating the dynamic complexity inherent in contemporary supply chain planning.

\subsubsection{Intent Classification}
%

Whereas the objective of the Query Enhancement Module in Section \ref{query_enhance_section} is to syntactically structure and parameterize the raw input, the focus here shifts to semantic interpretation, specifically the deciphering of the underlying business purpose driving that query.
The Intent Classification Agent functions as a specialized cognitive unit responsible for mapping these structured inputs onto well-defined supply chain planning contexts. Underpinned by a Large Language Model that has been fine-tuned on domain-specific corpora, this agent possesses the capability to navigate the linguistic nuances and contextual ambiguities inherent in operational directives. This domain adaptation is critical for overcoming the challenge of parsing vague or incomplete requests, where extracting the correct strategic signals is paramount. Guided by tailored prompt engineering, the agent systematically analyzes the user's input and categorizes the identified intent into three primary categories:
\begin{itemize}
    \item \textbf{Inventory Turnover and Planning Diagnostics} 
    This intent focuses on evaluating the efficiency of inventory movement and identifying potential issues in planning. Tasks under this intent may involve analyzing inventory turnover rates, detecting anomalies in replenishment schedules, and providing diagnostic insights for underperforming SKUs.
    The purpose of this classification is to enable proactive identification of inefficiencies, supporting adjustments that enhance supply chain responsiveness and reduce unnecessary inventory holding.
    
    \item \textbf{In-Stock Rate Monitoring}
    This intent pertains to monitoring product availability across warehouses or distribution centers. It includes tracking stock levels, flagging items at risk of stockouts, and alerting relevant agents for replenishment.
    High in-stock rates are crucial for maintaining service levels and customer satisfaction. Categorizing this intent allows the system to prioritize attention to critical items and high-demand products.

    \item \textbf{Sales, Inventory, and Procurement Recommendations}
    This intent aims to provide actionable guidance for ordering, procurement, and inventory allocation based on historical sales, current inventory, and demand forecasts. Sub-tasks include calculating optimal order quantities, recommending warehouse-specific allocations, and highlighting potential overstock or understock scenarios.
    Classifying this intent ensures that the system produces data-driven, actionable recommendations that align operational decisions with strategic planning objectives.
\end{itemize}




Subsequent to intention classification, the system directs the request to a specialized agent tailored to the specific business objective, including the Planning Generation Agent, Plan Diagnosis Agent, or Procurement–Inventory Recommendation Agent. Given that each domain entails distinct operational logic and idiosyncratic constraints, this routing mechanism enables the deployment of dedicated agents, each fortified with bespoke prompt engineering and fine-tuned models to ensure precise semantic alignment and execution.

Notwithstanding this functional divergence, the underlying cognitive architecture across these agents remains structurally isomorphic. All specialized agents operate upon a standardized workflow that encompasses task parsing, data retrieval, analytical reasoning, plan generation, and result validation. For the sake of conceptual clarity, we illustrate this unified orchestration paradigm in Figure 2, acknowledging that the detailed customization for each agent (e.g., task-specific prompts and constraint templates) corresponds to the distinct operational requirements discussed above.


This architectural configuration strikes a critical balance between modularity and specialization. Modularity enhances system scalability and maintainability, facilitating the seamless integration of novel task archetypes; conversely, specialization ensures rigorous domain alignment and decision accuracy for specific business contexts.
With the operational roles of these specialized agents defined, the subsequent analysis focuses on the Task Orchestration mechanism. This component governs how these distinct cognitive units are coordinated to collaboratively accomplish multi-step, inter-dependent planning objectives.

\subsubsection{Task Orchestration}
The user’s original text input functions as a conversational task, manifesting as a specific request or problem for which the user seeks system feedback and resolution. However, even after the system identifies the specific intent behind the input, the original task itself remains non-executable. Therefore, the next step is task orchestration, a process that breaks the request down into multiple system-executable subtasks. As detailed in the system prompt design (Figure \ref{prompt_planning}), this mechanism ensures that vague user inputs are translated into precise, actionable planning steps.

The \textbf{Task Orchestration Agent} is responsible for translating classified intents into structured, executable sub-tasks and coordinating their execution within the intelligent supply chain planning framework. Based on the output of the Intent Classification Agent, each user query is decomposed into sub-tasks that fall into one of three categories: \textit{historical sales data analysis}, \textit{feature data processing}, and \textit{plan formulation}.

\begin{itemize}
    \item \textbf{Historical Sales Data Analysis:} 
    Tasks in this category focus on analyzing past sales records to extract demand patterns, seasonal trends, and anomalies. This provides the quantitative foundation for subsequent planning decisions.

    \item \textbf{Feature Data Processing:} 
    These tasks involve transforming raw sales and inventory data into features suitable for predictive modeling or rule-based planning. Examples include generating aggregated statistics, calculating demand ratios, and normalizing data across product categories and warehouses.

    \item \textbf{Plan Formulation:} 
    This category includes generating actionable supply chain plans, such as order recommendations, inventory allocation, and replenishment schedules. It leverages insights from historical analysis and feature processing to ensure plans are both feasible and aligned with strategic objectives.
\end{itemize}

The orchestration process is iterative: after each sub-task is executed by the Task Execution Agent, results are returned to the orchestration module for evaluation and potential reconfiguration of remaining tasks. This loop continues until all sub-tasks are completed and a final, consolidated plan is produced.

By systematically decomposing and orchestrating tasks according to intent, the agent ensures that downstream execution is efficient, coherent, and adaptable. Moreover, the iterative feedback mechanism allows the framework to dynamically adjust task priorities and execution sequences based on intermediate results, enhancing robustness in complex and uncertain supply chain environments.

\begin{figure*}
    \centering
    \begin{tcolorbox}[colback=cyan!5!white,
      colframe=cyan!40!black,
      title=Prompt for Task Orchestration Agent]
    Answer the following questions as best you can. Use the following format: 
    
    Question: The input question you must answer. 
    
    Original task list: You should come up with a simple step-by-step plan
    
    Next task: the description of the task to solve in the next step
    
    Observation: the result of the task... (this updated task list/next task/Observation can be repeated zero or more times)
    
    Final Answer: the final answer to the original input question
    
    Begin!
    \end{tcolorbox}
    
    \caption{Prompt for Task Orchestration Agent}
    \label{prompt_planning}
\end{figure*}

\subsection{Task Execution Agents}

Following the successful decomposition of strategic intent by the Task Management module, the workflow advances to the Task Execution component. This module functions as the computational engine of the framework, responsible for materializing the orchestrated logical tasks into concrete data operations. 
Mirroring the architectural philosophy of the preceding section, this component employs a dual-agent structure comprising the Data Acquisition Agent and the Data Analysis Agent, both empowered by domain-adapted Large Language Models and specialized prompt engineering.

This execution phase bridges the critical gap between task planning and analytical realization. The Data Acquisition Agent first interfaces with heterogeneous data repositories to navigate the complexity of multi-source integration, ensuring the retrieval of clean, synchronized operational signals. Subsequently, the Data Analysis Agent processes this consolidated information to perform high-dimensional analytical synthesis, transforming raw data into actionable insights and forecasts. By segregating data retrieval from analytical reasoning, this modular design ensures that the execution process remains robust and rigorously aligned with strategic mandates, ultimately guaranteeing the data fidelity and analytical depth required for an adaptive and responsive supply chain.


\subsubsection{Data Acquisition}
This agent retrieves the required data according to task descriptions. It leverages \textit{Text-to-SQL} \parencite{zhu2024large} techniques to automatically translate natural language queries into SQL statements, ensuring accurate and efficient access to relevant historical and operational datasets. 

In the data acquisition stage, the primary goal of the system is to transform the intent-recognized natural language input into structured and executable parameters. Slot extraction plays a crucial role in this process, as it identifies the specific variables and feature names required for task execution. This step ensures that information can be accurately populated in subsequent function calls and database queries. Slot information typically includes key task elements such as the target object, time range, metric type, and constraint conditions, serving as a vital bridge between natural language understanding and programmatic execution.

At the implementation level, slot information is first vectorized using an embedding-based semantic model, which maps it into a high-dimensional semantic space. This vector representation effectively captures the semantic similarity among terms, enabling efficient retrieval and matching in later stages. The extracted slot vectors, slot values, and contextual information are stored in two key repositories: a dimensional vector database and an API database. The former supports similarity-based retrieval and slot matching—for instance, quickly associating the user input ``sales'' with the feature ``revenue.'' The latter stores standardized parameter and interface information to facilitate accurate function invocation and parameter acquisition.

To handle incomplete or ambiguous user inputs, a generative model is incorporated to assist in slot generation and completion. When certain slots are missing, unclear, or have low confidence, the generative model infers potential slot types and candidate values based on contextual semantics, enhancing both completeness and robustness of task execution. Through this mechanism, the slot extraction component establishes a stable mapping between semantic understanding and structured execution, enabling a seamless transition from natural language input to underlying data operations.


\subsubsection{Data Analysis}

\begin{figure*}
\begin{tcolorbox}[
    colback=cyan!5!white,
  colframe=cyan!40!black,
  title=Prompt for Task Execution Agent
]

You have access to a \texttt{pandas} DataFrame \texttt{df}: \\
\texttt{<dataframe description>}\\[1em]
Generate code to answer the following questions as best you can.\\[1em]
\textbf{Question:} the input question you must answer\\[1em]
Update this initial code:
\begin{verbatim}
# TODO: import the required dependencies
import pandas as pd

# Thought: explain what an Operation is needed, operation should be 
# one of [Filter Rows, Transform Columns, Groupby, Sort].
# Operation Code: Python code for this atomic operation. 
# Write code here

# ... (this Thought/Operation Code can repeat N times)
# Declare result var:
\end{verbatim}

Variable \texttt{df: pd.DataFrame} is already declared.\\
At the end, declare \texttt{result\_df} variable as the final DataFrame containing the answer.\\[1em]
\textbf{Begin!}\\
\textbf{Question:}
\end{tcolorbox}
\caption{Prompt for Task Execution Agent}
\label{prompt_execution}
\end{figure*}

The Data Analysis Agent operates as a specialized code synthesis engine, translating analytical objectives into executable scripts. To optimize for modularity and scalability, as codified in the system prompt shown in Figure \ref{prompt_execution}, we structure the code generation process around a grammar of four atomic primitive operations: \textit{Filter}, \textit{Transform}, \textit{Groupby}, and \textit{Sort}. Rather than attempting monolithic code generation, the system employs an iterative inference strategy: each reasoning cycle produces the code for a single atomic step, and these steps are progressively chained to construct a comprehensive, executable workflow. This granular architectural design confers several distinct advantages:

\begin{itemize}
    \item \textbf{Enhanced Automation Efficiency}: Decomposing complex analytical logic into discrete atomic primitives significantly streamlines the generative process. By restricting each inference cycle to a single operation, the system mitigates the model's cognitive load, thereby minimizing hallucination risks and optimizing both generation latency and execution reliability.

    \item \textbf{Enhanced Interpretability and Auditability}: The atomic design enforces a transparent, step-by-step data lineage. Since each operation possesses well-defined input-output contracts, the resulting workflow is inherently modular and verifiable. This explicit structural clarity not only streamlines fault isolation during debugging but also facilitates rigorous human-in-the-loop verification, ensuring that the analytical logic remains fully traceable and trustworthy.

    \item \textbf{Extensible Generalizability and Scalability}: The paradigm of atomic primitives transcends specific use cases, offering a composable logic that scales from routine data processing to intricate business simulations. By dynamically orchestrating these primitives, the framework accommodates diverse analytical requirements with high flexibility. Furthermore, this standardized interaction format facilitates the systematic accumulation of high-quality instruction-tuning corpora, establishing a rigorous data foundation that supports the continuous fine-tuning and self-evolution of the underlying large models.
\end{itemize}



Complementing the dynamic synthesis of atomic code, the Task Execution Agent leverages a robust \textit{function calling} mechanism to orchestrate encapsulated analytical modules. In scenarios demanding established mathematical precision, the central reasoning engine pivots from generating raw code to directly invoking pre-validated algorithmic tools via this mechanism. This capability enables the seamless integration of specialized domain assets, ranging from advanced predictive analytics engines (e.g., time-series forecasting) to complex Operations Research solvers (e.g., inventory optimization). By strategically offloading computation-intensive tasks to these deterministic engines, the system ensures that critical decision outputs maintain strict mathematical rigor and business consistency, thereby liberating the foundation model to concentrate on the high-level orchestration of the overall analytical pipeline.

Ultimately, the Task Execution module integrates these capabilities into a unified, iterative workflow. Through a sequential reasoning process, the agent dynamically alternates between generating atomic code segments and invoking specialized domain tools. This stepwise construction allows the system to progressively assemble a complete executable implementation, effectively combining the flexibility of generative logic with the precision of established analytical models.





\subsection{Iterative Execution and Dynamic Subtask Refinement}

To navigate the stochasticity of supply chain scenarios, the framework implements an iterative execution strategy centered on the dynamic refinement of the \textit{decomposed subtask list}. Rather than rigidly processing the initial decomposition sequence, the system treats the task queue as a fluid, state-dependent entity. Upon the completion of each individual subtask, the Planning Agent consolidates the execution feedback to reassess the validity of the remaining items in the list. This triggers a cycle of dynamic list maintenance—where pending subtasks are systematically pruned, re-sequenced, or augmented based on real-time intermediate outcomes. This recursive \textit{execute–evaluate–refine} loop ensures that the overarching plan is not static, but continuously reconstructs itself to align with the evolving operational reality.

\subsection{Plan Correction Agent}
After a sales plan is formulated and implemented, it is essential to monitor the plan’s execution in real time. Specifically, the actual daily sales and other relevant metrics are compared against the planned values. Based on the progress toward targets, the plan may require timely adjustments and updates to ensure alignment with business goals.

The Plan Correction Agent assists business personnel by automatically analyzing whether the plan execution is on track.
It detects abnormal deviations from expected progress and identifies underlying causes, ranging from lower-than-expected traffic and underperforming promotions to unexpected market shifts, before generating recommendations for plan updates.

By integrating continuous observation, data analysis, and actionable recommendations, the Plan Correction Agent enables a dynamic feedback loop, ensuring that the sales plan remains adaptive and responsive to real-world conditions. This mechanism significantly improves the robustness and effectiveness of supply chain planning, particularly in fast-changing e-commerce environments. 


\section{From Design to Deployment}

To validate the practical efficacy of the proposed framework, we conduct a comprehensive empirical study within the JD.com supply chain network. This environment serves as a representative testbed characterized by significant operational scale and heterogeneity. The platform manages a portfolio of approximately 10 million SKUs and serves a network of 8 million enterprises. It encompasses diverse supply chain modalities that range from high-turnover consumer goods to long-tail durable products. This operational complexity provides a rigorous scenario for testing the adaptability and robustness of intelligent planning methods under real-world constraints.

Since the initial deployment in December 2023, the framework has undergone continuous iterative refinement within the live production environment. For this empirical evaluation, we specifically target a representative portfolio of over \textbf{70,000} distinct SKUs across three core business units: Grain \& Oils, Maternal \& Infant Care, and Small Home Appliances. This selection explicitly targets distinct inventory behaviors, allowing us to evaluate the framework's capability to generalize across divergent demand patterns.

\textbf{Process Efficiency: Workflow Reconfiguration.}
The deployment drives a fundamental leap in workforce efficiency. For an individual planner within each business unit, the manual baseline imposes \textbf{a weekly cost of 120 minutes}. This breakdown comprises 20 minutes of acquisition, 40 minutes of processing, and 60 minutes of analysis. The framework streamlines this time-intensive process. The user provides a data requirement description in just 5 minutes. The system fully automates the processing phase and reduces the manual effort for this step to zero. Consequently, the analytical hour evolves into an interactive, agent-supported session. This shift liberates human cognition from routine data preparation to focus entirely on strategic reasoning.

\textbf{Decision Quality: Standardization and Precision.}
The framework elevates the baseline of planning accuracy by standardizing the decision workflow. It codifies the rigorous data processing and decision logic of veteran planners into automated protocols. This decouples decision quality from individual variances and ensures consistent inputs across the organization.

To quantify this impact, the study leverages a Year-over-Year comparison between May–June 2024 (Agent-driven) and May–June 2023 (Manual baseline). This window encompasses the critical ``618'' Shopping Festival, which imposes distinct operational requirements: a Stock-up Phase for accumulation and a Sales Peak Phase for rapid depletion. Within this dynamic context, the multi-agent system orchestrates the deployment of inventory decision algorithms to adapt to these shifting objectives. Under a rigorous metric defining deviation based on end-of-week inventory values, the system achieves a \textbf{22\%} increase in the proportion of plans with an accuracy deviation below \textbf{5\%}. This confirms the framework's ability to operationalize complex decision logic across different phases.


\textbf{Operational Resilience: Closed-Loop Execution and Value Realization.}
The framework demonstrates robust operational resilience under conditions of high demand uncertainty. Empirical evidence from the ``618'' Shopping Festival confirms this capability. We define the stock fulfillment rate at the granular SKU-Distribution Center (SKU-DC) level. It measures the daily proportion of SKU-DC pairs that maintain a positive end-of-day inventory balance relative to the total active portfolio. Despite peak volatility, the system maintains a stock fulfillment rate \textbf{2\%} above the historical manual baseline. This metric indicates a substantial improvement in service levels and confirms the system's ability to mitigate supply-demand mismatches in a high-velocity retail environment.

This performance gain stems from a fundamental reduction in decision latency. Unlike manual workflows constrained by periodic batch reviews, the multi-agent system operates on a continuous, event-driven loop. By instantly converting demand signals into execution orders, the framework minimizes stockout durations and captures fleeting sales opportunities. Consequently, we estimate that this agility translates to a Gross Merchandise Volume (GMV) uplift of approximately \textbf{RMB 2 million} during the festival.

Collectively, the field deployment serves as the empirical embodiment of our architectural rethinking. It translates the proposed agentic paradigm into a tangible operational reality within a complex industrial environment. The demonstrated improvements across process efficiency, decision quality, and operational resilience substantiate the validity of this theoretical shift. 

\section{Conclusion}

In this paper, we redefine supply chain planning as \textit{an interactive, integrated, and automated generative process that systematically produces vertically coherent, horizontally synergistic, and dynamically adaptive plans}. To derive this conceptualization, we synthesize the historical evolution of the discipline with an analysis of converging internal and external drivers. This examination reveals that legacy stability assumptions are untenable in next-generation supply chain environments. Consequently, we identify Generative AI as the essential enabler that provides the semantic reasoning capabilities necessary to operationalize this framework, driving a fundamental shift from rigid structure to adaptive co-evolution.

We materialize this paradigm by developing a Generative AI-powered agentic framework that functions as an intelligent organizational interface. This architecture orchestrates the planning lifecycle through integrated Task Management and Task Execution modules. Specialized agents collaborate to decompose strategic intent into executable code, supported by an iterative feedback mechanism for continuous alignment. The technical architecture of this workflow entails complexities that extend far beyond basic prompt engineering. Critical imperatives include the design of robust databases for dynamic retrieval, precise semantic understanding of user intent, and the rigorous transformation of unstructured inputs into structured formats—challenges that must be addressed to enable scalable, end-to-end intelligent decision-making.

We assess the practical utility of this framework through a field study within JD.com's supply chain network. Regarding efficiency, the automation of routine data processing reduces the weekly manual workload for planners from 120 minutes to approximately 5 minutes. In terms of decision quality, the standardized reasoning logic results in a 22\% increase in the proportion of plans with high accuracy. Furthermore, in high-volatility scenarios such as the ``618'' Shopping Festival, the system maintains a stock fulfillment rate 2\% above the historical baseline. These findings indicate that Generative AI functions effectively as an operational agent in complex stochastic environments.

Overall, we position supply chain planning as the strategic beachhead for the discipline's reconstruction. By inherently bridging strategic intent with operational execution, planning aligns naturally with Generative AI. We recognize that substantial potential remains to further advance this paradigm. Specifically, we advocate for developing reasoning-enhanced foundation models, integrating deep domain knowledge, and expanding algorithmic tool capabilities. Ultimately, we envision these principles extending beyond the planning domain to inform a holistic transformation across the full spectrum of supply chain management, from intelligent sourcing to resilient logistics. This evolution paves the way for a future where supply chains are not merely managed, but autonomously orchestrated.

\printbibliography

@incollection{chopra2019supply,
  title={Supply chain management. Strategy, planning \& operation},
  author={Chopra, Sunil and Meindl, Peter},
  booktitle={Das Summa Summarum des Management: Die 25 wichtigsten Werke f{\"u}r Strategie, F{\"u}hrung und Ver{\"a}nderung},
  pages={265--275},
  year={2019},
  publisher={Springer}
}

@book{stadtler2015supply,
  title={Supply chain management and advanced planning: concepts, models, software, and case studies},
  author={Stadtler, Hartmut and Stadtler, Hartmut and Kilger, Christoph and Kilger, Christoph and Meyr, Herbert and Meyr, Herbert},
  year={2015},
  publisher={Springer}
}

@article{srivastava2024exploring,
  title={Exploring the potential of large language models in supply chain management: A study using big data},
  author={Srivastava, Santosh Kumar and Routray, Susmi and Bag, Surajit and Gupta, Shivam and Zhang, Justin Zuopeng},
  journal={Journal of Global Information Management (JGIM)},
  volume={32},
  number={1},
  pages={1--29},
  year={2024},
  publisher={IGI Global Scientific Publishing}
}

@article{quan2024invagent,
  title={Invagent: A large language model based multi-agent system for inventory management in supply chains},
  author={Quan, Yinzhu and Liu, Zefang},
  journal={arXiv preprint arXiv:2407.11384},
  year={2024}
}

@article{jannelli2024agentic,
  title={Agentic LLMs in the Supply Chain: Towards Autonomous Multi-Agent Consensus-Seeking},
  author={Jannelli, Valeria and Schoepf, Stefan and Bickel, Matthias and Netland, Torbj{\o}rn and Brintrup, Alexandra},
  journal={arXiv preprint arXiv:2411.10184},
  year={2024}
}

@article{li2023large,
  title={Large language models for supply chain optimization},
  author={Li, Beibin and Mellou, Konstantina and Zhang, Bo and Pathuri, Jeevan and Menache, Ishai},
  journal={arXiv preprint arXiv:2307.03875},
  year={2023}
}

@article{zhu2024large,
  title={Large language model enhanced text-to-sql generation: A survey},
  author={Zhu, Xiaohu and Li, Qian and Cui, Lizhen and Liu, Yongkang},
  journal={arXiv preprint arXiv:2410.06011},
  year={2024}
}

@article{heiney2021intel,
  title={Intel realizes \$25 billion by applying advanced analytics from product architecture design through supply chain planning},
  author={Heiney, John and Lovrien, Ryan and Mason, Nicholas and Ovacik, Irfan and Rash, Evan and Sarkar, Nandini and Travis, Harry and Zhao, Zhenying and Ching, Kalani and Shirodkar, Shamin and others},
  journal={INFORMS Journal on Applied Analytics},
  volume={51},
  number={1},
  pages={9--25},
  year={2021},
  publisher={INFORMS}
}

@article{lee2021new,
  title={The new AAA supply chain},
  author={Lee, Hau L},
  journal={Management and Business Review},
  volume={1},
  number={1},
  pages={173--176},
  year={2021},
  publisher={SAGE Publications Sage CA: Los Angeles, CA}
}

@book{orlicky1974material,
  title={Material requirements planning: the new way of life in production and inventory management},
  author={Orlicky, Joseph A},
  year={1974},
  publisher={McGraw-Hill, Inc.}
}

@article{hax1973hierarchical,
  title={Hierarchical integration of production planning and scheduling},
  author={Hax, Arnoldo C and Meal, Harlan C},
  year={1973},
  publisher={MIT]}
}

@article{bitran1993hierarchical,
  title={Hierarchical production planning},
  author={Bitran, Gabriel R and Tirupati, Devanath},
  journal={Handbooks in operations research and management science},
  volume={4},
  pages={523--568},
  year={1993},
  publisher={Elsevier}
}

@article{lee1992managing,
  title={Managing supply chain inventory: pitfalls and opportunities},
  author={Lee, Hau L and Billington, Corey},
  journal={Sloan management review},
  volume={33},
  number={3},
  pages={65--73},
  year={1992}
}

@article{lee1993material,
  title={Material management in decentralized supply chains},
  author={Lee, Hau L and Billington, Corey},
  journal={Operations research},
  volume={41},
  number={5},
  pages={835--847},
  year={1993},
  publisher={INFORMS}
}

@book{simchi1999designing,
  title={Designing and managing the supply chain: Concepts, strategies, and cases},
  author={Simchi-Levi, David and Kaminsky, Philip and Simchi-Levi, Edith},
  year={1999},
  publisher={McGraw-hill New York}
}

@article{lee2004triple,
  title={The triple-A supply chain},
  author={Lee, Hau L and others},
  journal={Harvard business review},
  volume={82},
  number={10},
  pages={102--113},
  year={2004}
}

@article{ivanov2020viability,
  title={Viability of intertwined supply networks: extending the supply chain resilience angles towards survivability. A position paper motivated by COVID-19 outbreak},
  author={Ivanov, Dmitry and Dolgui, Alexandre},
  journal={International Journal of Production Research},
  volume={58},
  number={10},
  pages={2904--2915},
  year={2020},
  publisher={Taylor \& Francis}
}

@article{zhang2022green,
  title={Green supply chain integration, supply chain agility and green innovation performance: Evidence from Chinese manufacturing enterprises},
  author={Zhang, Bochen and Zhao, Shukuan and Fan, Xueyuan and Wang, Shuang and Shao, Dong},
  journal={Frontiers in Environmental Science},
  volume={10},
  pages={1045414},
  year={2022},
  publisher={Frontiers Media SA}
}

@article{tarigan2021impact,
  title={Impact of internal integration, supply chain partnership, supply chain agility, and supply chain resilience on sustainable advantage},
  author={Tarigan, Zeplin Jiwa Husada and Siagian, Hotlan and Jie, Ferry},
  journal={Sustainability},
  volume={13},
  number={10},
  pages={5460},
  year={2021},
  publisher={MDPI}
}

@article{shen2023strengthening,
  title={Strengthening supply chain resilience during COVID-19: A case study of JD. com},
  author={Shen, Zuojun Max and Sun, Yiqi},
  journal={Journal of operations management},
  volume={69},
  number={3},
  pages={359--383},
  year={2023},
  publisher={Wiley Online Library}
}

@article{xu2024towards,
  title={Towards autonomous supply chains: Definition, characteristics, conceptual framework, and autonomy levels},
  author={Xu, Liming and Mak, Stephen and Proselkov, Yaniv and Brintrup, Alexandra},
  journal={Journal of Industrial Information Integration},
  volume={42},
  pages={100698},
  year={2024},
  publisher={Elsevier}
}

@article{raj2025advancing,
  title={Advancing supply chain management from agility to hyperagility: a dynamic capability view},
  author={Raj, Alok and Sharma, Varun and Shukla, Dhirendra Mani and Sharma, Prateek},
  journal={Annals of Operations Research},
  volume={348},
  number={3},
  pages={1457--1488},
  year={2025},
  publisher={Springer}
}

@article{koneti2024human,
  title={Human-Machine Collaboration in Supply Chain Management: The Impact of AI on Workforce Dynamics},
  author={Koneti, C and Sajja, GS and Adarsh, A and Yerasuri, SS and Mann, G and Mandal, A},
  journal={Journal of Informatics Education and Research},
  volume={4},
  number={3},
  pages={934--943},
  year={2024}
}

@article{aljohani2023predictive,
  title={Predictive analytics and machine learning for real-time supply chain risk mitigation and agility},
  author={Aljohani, Abeer},
  journal={Sustainability},
  volume={15},
  number={20},
  pages={15088},
  year={2023},
  publisher={MDPI}
}

@article{jonsson2013centralised,
  title={Centralised supply chain planning at IKEA},
  author={Jonsson, Patrik and Rudberg, Martin and Holmberg, Stefan},
  journal={Supply Chain Management: An International Journal},
  volume={18},
  number={3},
  pages={337--350},
  year={2013},
  publisher={Emerald Group Publishing Limited}
}

@article{herden2020explaining,
  title={Explaining the competitive advantage generated from Analytics with the knowledge-based view: the example of Logistics and Supply Chain Management},
  author={Herden, Tino T},
  journal={Business Research},
  volume={13},
  number={1},
  pages={163--214},
  year={2020},
  publisher={Springer}
}

@article{ngo2023digital,
  title={Digital supply chain transformation: effect of firm’s knowledge creation capabilities under COVID-19 supply chain disruption risk},
  author={Ngo, Vu Minh and Nguyen, Huan Huu and Pham, Hiep Cong and Nguyen, Hung Manh and Truong, Phuc Vinh Dang},
  journal={Operations Management Research},
  volume={16},
  number={2},
  pages={1003--1018},
  year={2023},
  publisher={Springer}
}

@article{tang2019strategic,
  title={The strategic role of logistics in the industry 4.0 era},
  author={Tang, Christopher S and Veelenturf, Lucas P},
  journal={Transportation Research Part E: Logistics and Transportation Review},
  volume={129},
  pages={1--11},
  year={2019},
  publisher={Elsevier}
}

@article{shen2025jd,
  title={JD. com Improves Fulfillment Efficiency with Data-Driven Integrated Assortment Planning and Inventory Allocation},
  author={Shen, Zuo-Jun Max and Sun, Shuo and Qi, Yongzhi and Hu, Hao and Kang, Ningxuan and Zhang, Jianshen and Wang, Xin and Lin, Xiaoming},
  journal={INFORMS Journal on Applied Analytics},
  volume={55},
  number={5},
  pages={386--398},
  year={2025},
  publisher={INFORMS}
}

@article{hu2024supercharged,
  title={Supercharged by advanced analytics, jd. com attains agility, resilience, and shared value across its supply chain},
  author={Hu, Hao and Qi, Yongzhi and Lee, Hau L and Shen, Zuo-Jun Max and Liu, Curtis and Zhu, Weimeng and Kang, Ningxuan},
  journal={INFORMS Journal on Applied Analytics},
  volume={54},
  number={1},
  pages={54--70},
  year={2024},
  publisher={INFORMS}
}

@article{zhang2025unstructured,
  title={From unstructured communication to intelligent RAG: Multi-agent automation for supply chain knowledge bases},
  author={Zhang, Yao and Shang, Zaixi and Patel, Silpan and Zuniga, Mikel},
  journal={arXiv preprint arXiv:2506.17484},
  year={2025}
}

@techreport{syntun2024double11,
  title        = {2024 {Double 11 Shopping Festival} Report},
  author       = {{Syntun}},
  year         = {2024},
  institution  = {Syntun Data Technology Co., Ltd.},
  address      = {Beijing, China},
  url          = {https://www.syntun.com/en/news/2024-double11-report},
  note         = {Accessed: 2025-11-13},
}

@article{garcia2023creating,
  title={Creating and sharing interorganizational knowledge through a supply chain 4.0 project: a case study},
  author={Garcia, Fabienne and Grabot, Bernard and Pach{\'e}, Gilles},
  journal={Journal of Global Information Management (JGIM)},
  volume={31},
  number={1},
  pages={1--19},
  year={2023},
  publisher={IGI Global}
}

@misc{EU_ETS,
  author = {{European Commission}},
  title = {EU Emissions Trading System (EU ETS)},
  year = {2025},
  howpublished = {\url{https://climate.ec.europa.eu/eu-action/carbon-markets/eu-emissions-trading-system-eu-ets_en}}
}

@misc{China_Carbon,
  author = {{State Council of China}},
  title = {China's Carbon Peak and Carbon Neutrality Action Plan},
  year = {2025},
  howpublished = {\url{https://english.www.gov.cn/archive/whitepaper/202511/08/content_WS690ee812c6d00ca5f9a076cd.html}}
}

@misc{WorldBank_CarbonPricing,
  author = {{World Bank}},
  title = {State and Trends of Carbon Pricing 2021},
  year = {2021},
  howpublished = {\url{https://openknowledge.worldbank.org/handle/10986/35620}}
}

@article{prakash2025generative,
  title={Generative AI in Supply Chain: A Systematic Review of Opportunities, Benefits, and Challenges},
  author={Prakash, Chandra and Shaikh, Musa and Mutha, Pavan},
  journal={Asian Journal of Research in Computer Science},
  volume={18},
  number={10},
  pages={170--182},
  year={2025}
}

@article{riad2024enhancing,
  title={Enhancing supply chain resilience through artificial intelligence: developing a comprehensive conceptual framework for AI implementation and supply chain optimization},
  author={Riad, Meriem and Naimi, Mohamed and Okar, Chafik},
  journal={Logistics},
  volume={8},
  number={4},
  pages={111},
  year={2024},
  publisher={MDPI}
}

@article{teixeira2025intelligent,
  title={Intelligent supply chain management: A systematic literature review on artificial intelligence contributions},
  author={Teixeira, Ant{\'o}nio R and Ferreira, Jos{\'e} Vasconcelos and Ramos, Ana Lu{\'\i}sa},
  journal={Information},
  volume={16},
  number={5},
  pages={399},
  year={2025},
  publisher={MDPI}
}

@article{venkatachalam2025integrating,
  title={Integrating Large Language Models with Network Optimization for Interactive and Explainable Supply Chain Planning: A Real-World Case Study},
  author={Venkatachalam, Saravanan},
  journal={arXiv preprint arXiv:2508.21622},
  year={2025}
}

@article{daios2025ai,
  title={AI applications in supply chain management: A survey},
  author={Daios, Adamos and Kladovasilakis, Nikolaos and Kelemis, Athanasios and Kostavelis, Ioannis},
  journal={Applied Sciences},
  volume={15},
  number={5},
  pages={2775},
  year={2025},
  publisher={MDPI}
}

@article{aftab2018super,
  title={Super responsive supply chain: The case of Spanish fast fashion retailer Inditex-Zara},
  author={Aftab, Md Afzalul and Yuanjian, Qin and Kabir, Nadia and Barua, Zapan},
  journal={International Journal of Business and Management},
  volume={13},
  number={5},
  pages={212--227},
  year={2018},
  publisher={Canadian Center of Science and Education}
}

@misc{scmglobe2024,
  author = {{SCM Globe}},
  title = {Zara Clothing Company Supply Chain},
  year = {2024},
  url = {https://www.scmglobe.com/zara-clothing-company-supply-chain/},
}

@misc{walmart2018,
  author = {{Walmart Newsroom}},
  title = {Unilever and Walmart announce forest sustainability initiatives at the Global Climate Action Summit},
  year = {2018},
  url = {https://corporate.walmart.com/news/2018/09/13/unilever-and-walmart-announce-forest-sustainability-initiatives-at-the-global-climate-action-summit},
}

@misc{aws2023,
  author = {{AWS Supply Chain Blog}},
  title = {Improve supply planning accuracy with machine-learning-based lead time insights},
  year = {2023},
  url = {https://aws.amazon.com/blogs/supply-chain/improve-supply-planning-accuracy-with-machine-learning-based-lead-time-insights/},
}

@article{simchi2025large,
  title={Large language models for supply chain decisions},
  author={Simchi-Levi, David and Mellou, Konstantina and Menache, Ishai and Pathuri, Jeevan},
  journal={arXiv preprint arXiv:2507.21502},
  year={2025}
}

@article{brown2020language,
  title={Language models are few-shot learners},
  author={Brown, Tom and Mann, Benjamin and Ryder, Nick and Subbiah, Melanie and Kaplan, Jared D and Dhariwal, Prafulla and Neelakantan, Arvind and Shyam, Pranav and Sastry, Girish and Askell, Amanda and others},
  journal={Advances in neural information processing systems},
  volume={33},
  pages={1877--1901},
  year={2020}
}

@article{achiam2023gpt,
  title={Gpt-4 technical report},
  author={Achiam, Josh and Adler, Steven and Agarwal, Sandhini and Ahmad, Lama and Akkaya, Ilge and Aleman, Florencia Leoni and Almeida, Diogo and Altenschmidt, Janko and Altman, Sam and Anadkat, Shyamal and others},
  journal={arXiv preprint arXiv:2303.08774},
  year={2023}
}

@article{simchi2014superstorms,
  title={From superstorms to factory fires: Managing unpredictable supply chain disruptions},
  author={Simchi-Levi, David and Schmidt, William and Wei, Yehua},
  journal={Harvard Business Review},
  volume={92},
  number={1-2},
  pages={96--101},
  year={2014}
}

\clearpage
\begin{APPENDIX}{Detailed Workflow and Input-Output Specifications}

Our system can be regarded as a conversational assistant designed for business personnel, where the tasks it handles are inherently diverse. To provide a more intuitive understanding of each step’s specific content, as well as the corresponding inputs and outputs, we present a simple illustrative example. Consider a user request: \textit{``Generate the November sales plan for the Computer Department.''}

\noindent \textbf{Step 0: Retrieval of Standard Operating Procedure (SOP).} 

The agent first queries the organizational knowledge base via a retrieval-augmented generation (RAG) mechanism. 
The retrieved content corresponds to the department’s established SOPs for sales plan formulation. 
For the computer department, the SOP contains best-practice guidelines accumulated from prior planning processes, including:
\begin{enumerate}
    \item Analyzing historical sales, traffic, and user growth trends from both the same period in previous years and recent months.
    \item Identifying key factors influencing sales (e.g., seasonality, promotions, or external events).
    \item Formulating the sales plan for the upcoming period.
\end{enumerate}

\noindent \textbf{Step 1: Intent Classification.}

Given the current user request in the example, the first step of the framework is intent recognition, which aims to determine the purpose behind the input. In this particular case, the system identifies that the user’s request corresponds to plan generation.

This module employs a combination of the LLM and domain-specific heuristics to capture both the semantic content and contextual cues of the user's input. By accurately classifying the intent, the framework ensures that the subsequent plan generation module receives precise instructions, enabling the creation of a coherent and actionable plan aligned with the user's objectives.

\noindent \textbf{Step 2: Task Orchestration.}

The retrieved SOP is dynamically injected into a prompt template, replacing the placeholder variable \texttt{\{SOP RAG\}}. 
The Task Planning Agent then leverages the LLM to interpret the user request in light of the SOP and generate a structured task list. 
For the current example, the agent may output the following sub-tasks:

Through this mechanism, the Task Planning Agent aligns the user’s high-level intent with organizational best practices, ensuring that the resulting task list is both \textit{contextually relevant} and \textit{operationally grounded}.

\begin{tcolorbox}[
  colback=green!5!white,    
  colframe=green!40!black,  
  colbacktitle=white,       
  coltitle=black,           
  sharp corners,
  title=Output of Step 2: Generated Task List
]
\textbf{Original task list:}
\begin{enumerate}
    \item Analyze the computer department's historical and recent sales, traffic, and user growth, and identify the key factors influencing its performance.
    \item Analyze historical and recent sales, traffic, and user growth of the overall market and industry, and identify the key factors affecting sales.
    \item Formulate the computer department's November sales plan and adjust it based on external information.
\end{enumerate}

\textbf{Next task:} Analyze the computer department's historical and recent sales, traffic, and user growth, and identify the key factors influencing its performance.
\end{tcolorbox}
Once the task list is generated by the Task Planning Agent, each sub-task is executed by the Task Execution Agent. 
The execution process consists of multiple specialized agents that handle data retrieval, analysis, and subsequent planning steps. 

\noindent \textbf{Step 3: Data Retrieval by the Data Acquisition Agent}

The agent generates SQL fragments using text-to-SQL methods, assembles them into a full query, and executes it against the organizational database.
\begin{tcolorbox}[
  colback=green!5!white,    
  colframe=green!40!black,  
  colbacktitle=white,       
  coltitle=black,           
  sharp corners,
  title=Data Acquisition Agent: Input and Output
]
\textbf{Input:} Analyze the computer department's historical and recent sales, traffic, and user growth, and identify the key factors influencing its performance. \\[2mm]

\textbf{Output (SQL query):}
\begin{verbatim}
SELECT year, month, SUM(sales), 
SUM(pv), SUM(user_cnt)
FROM sales
WHERE order_date >= sysdate(-730) 
AND dept_id = 'Computer'
GROUP BY year, month
\end{verbatim}

\textbf{Output (Structured Data):} Retrieved data is organized into a \texttt{pandas DataFrame} for downstream analysis.
\end{tcolorbox}

\vspace{1em}

\noindent \textbf{Step 4: Data Analysis by the Data Analysis Agent}

 The agent generates Python scripts using \texttt{pandas} and other analytical tools to examine trends, seasonal patterns, and influencing factors.
\begin{tcolorbox}[
  colback=green!5!white,    
  colframe=green!40!black,  
  colbacktitle=white,       
  coltitle=black,           
  sharp corners,
  title=Data Analysis Agent: Input and Output
]
\textbf{Input:} Structured dataset of the computer department's historical sales, traffic, and user growth. Task: Analyze past and recent sales trends, traffic patterns, user growth, and identify key sales factors. \\[2mm]

\textbf{Output (Analysis Results):}
\begin{itemize}
    \item Sales Trends: The computer department’s November sales show stable growth over the past years, with peaks during holiday periods.
    \item Traffic: Recent page views show a slight decline, but overall user engagement remains high.
    \item User Growth: New user acquisition has slowed, while retention of existing users remains strong.
    \item Key Sales Drivers: Holiday promotions, product update frequency, competitor activity, marketing investments.
\end{itemize}
\end{tcolorbox}

This structured execution ensures that raw data is systematically transformed into actionable insights, providing a solid foundation for subsequent planning and forecasting tasks within the LLM Agent framework.

\noindent \textbf{Step 5: Task update}

After each sub-task is executed by the Task Execution Agent, the results are collected as observations. These observations capture the outputs of data retrieval and analysis, such as sales trends, traffic patterns, user growth metrics, and identified key factors affecting sales performance.

The Task Planning Agent then integrates these observations into its planning process. Specifically, the agent injects the new information into its prompt template, replacing placeholders like {Observation}, and re-evaluates the current task list. Completed tasks are pruned, and the agent may adjust the order of remaining tasks or add new subtasks based on the insights gained. The updated task list is then generated, highlighting the next sub-task to be executed. 

Through this planning–execution–observation–replanning loop, the Task Planning Agent continuously refines the task sequence. This mechanism enables the agent to maintain adaptability, ensure alignment with user goals, and respond effectively to real-time observations in complex and dynamic supply chain scenarios.

\begin{figure*}[t]
\centering
\begin{tcolorbox}[
  colback=green!5!white,    
  colframe=green!40!black,  
  colbacktitle=white,       
  coltitle=black,           
  sharp corners,
  title=Task Update: Input and Output
]

\textbf{Input:} 
\begin{itemize}
    \item \textbf{Question:} Generate the November sales plan for the computer department.
    \item \textbf{Original Task List:}
    \begin{enumerate}
        \item Analyze the computer department's historical and recent sales, traffic, and user growth, identifying key factors affecting its performance.
        \item Analyze market and industry historical and recent sales, traffic, and user growth, identifying factors influencing overall sales.
        \item Formulate the computer department's November sales plan, adjusting according to external information.
    \end{enumerate}
    \item \textbf{Next Task:} Analyze the computer department's historical and recent sales, traffic, and user growth, identifying key factors affecting its performance.
    \item \textbf{Observation:} the computer department's November sales over the past several years have shown stable growth, with pronounced peaks around holiday periods. Traffic data indicates a slight decline in recent visits, while user engagement remains high. User growth has slowed for new users, but retention of existing users remains strong. Key factors influencing sales include holiday promotions, product update frequency, competitor activity, and marketing investments.
\end{itemize}
\vspace{1em}
\textbf{Output: Updated Task List}
\begin{itemize}
    \item \textbf{Updated Task List}
    \begin{enumerate}
    \item Analyze market and industry historical and recent sales, traffic, and user growth, identifying factors influencing overall sales.
    \item Formulate the computer department's November sales plan, adjusting according to external information.
    \end{enumerate}
    \item \textbf{Next Task:} Analyze market and industry historical and recent sales, traffic, and user growth, identifying factors influencing overall sales.
\end{itemize}
\end{tcolorbox}
\end{figure*}

\textbf{Repeat Task Execution and Task List Update}s
After the Task Planning Agent updates the task list in Step 3, the process enters a loop of execution and planning. The next sub-task in the updated list is picked by the Task Execution Agent, and the cycle of data retrieval, analysis, and generation of observations is repeated.

Specifically, the Task Execution Agent sequentially executes the next sub-task, producing new observations, which are then fed back to the Task Planning Agent. The agent integrates these observations into the prompt template, dynamically adjusts the task list by pruning completed tasks and reordering or adding new tasks if needed, and outputs the next sub-task to execute.

This iterative cycle continues until all tasks in the task list are completed. The mechanism allows the agentic framework to adapt to newly acquired information, handle complex dependencies between subtasks, and maintain alignment with the user's high-level goals.

By repeatedly performing Steps 2–4, the system ensures that each task is executed in a context-aware manner and that the task list continuously evolves based on real-time insights, improving the robustness and accuracy of the supply chain planning process.

\begin{figure*}
\begin{tcolorbox}[
  colback=green!5!white,    
  colframe=green!40!black,  
  colbacktitle=white,       
  coltitle=black,           
  sharp corners,
  title= Final Task Execution Output
]

\textbf{Input:} 
\begin{itemize}
    \item \textbf{Updated Task List:}
    \begin{enumerate}
        \item Formulate the computer department's November sales plan, adjusting according to external information.
    \end{enumerate}
    \item \textbf{Next Task:} Formulate the computer department's November sales plan, adjusting according to external information.
    \item \textbf{Observation:} Based on detailed data analysis, the November sales plan for the computer department is as follows:
    \begin{itemize}
        \item Holiday Promotions: During important November holidays (e.g., Singles' Day), increase the promotional budget by 150\% relative to the baseline. Implement discounts, gifts, and other incentives to attract consumers. This is expected to increase sales by approximately 15\%.
    \end{itemize}
\end{itemize}

\vspace{1em}

\textbf{Final Output:} For November, the Computer Department should target a monthly sales goal of approximately RMB 120 million. This represents a 15\% growth over the previous period and can be achieved by increasing the holiday promotion budget by 150\%.
\end{tcolorbox}
\end{figure*}

\end{APPENDIX}

\end{document}